%% file: iclr2026_conference.tex
\documentclass{article} % For LaTeX2e
\usepackage{iclr2026_conference, times}

\input{math_commands}

\usepackage{hyperref}
\usepackage{url}
\usepackage{graphicx}
\graphicspath{{figures/}}
\usepackage{listings}

\usepackage[T1]{fontenc}
\usepackage[scaled]{inconsolata} % nicer mono (optional)
\usepackage[dvipsnames]{xcolor}
\usepackage{caption}
\usepackage{calc}      % for \numexpr
\usepackage{etoolbox}  % for robust macros
\usepackage{pythonhighlight}
\usepackage{multirow}

\usepackage{tabularx}
\usepackage{booktabs}

\newcolumntype{L}{>{\raggedright\arraybackslash}X}

\definecolor{pybg}{RGB}{246,246,242}
\definecolor{pycm}{RGB}{80,120,80}
\definecolor{pykw}{RGB}{0,70,173}
\definecolor{pystr}{RGB}{163,21,21}
\definecolor{pylineno}{gray}{0.45}

\lstdefinestyle{pyhex}{
  language=Python,
  backgroundcolor=\color{pybg},
  basicstyle=\ttfamily\footnotesize,
  numbers=left,
  numberstyle=\scriptsize\color{pylineno},
  numbersep=6pt,
  keywordstyle=\bfseries\color{pykw},
  commentstyle=\itshape\color{pycm},
  stringstyle=\color{pystr},
  showstringspaces=false,
  tabsize=2,
  keepspaces=true,
  breaklines=true,
  breakatwhitespace=true,
  columns=fullflexible,
  frame=single,
  framerule=0.2pt,
  rulecolor=\color{black!15},
  upquote=true,
  postbreak=\mbox{\textcolor{pylineno}{$\hookrightarrow$}\space},
}
\lstdefinestyle{pyhex-appendix}{
  style=pyhex,
  basicstyle=\ttfamily\scriptsize, % use \tiny to shrink further
  numberstyle=\tiny\color{pylineno},
}
\newcommand{\CodeFigureTwoCols}[5][0.9]{%
  \begin{figure}[t]
    \centering
    \begingroup
      \newcount\hex@split \newcount\hex@start
      \hex@split=#3\relax
      \hex@start=\hex@split \advance\hex@start by 1
      \scalebox{#1}{%
        \begin{minipage}{0.98\linewidth}
          \begin{minipage}[t]{0.49\linewidth}
            \lstinputlisting[style=pyhex-appendix,
              firstline=1,
              lastline=\number\hex@split]{#2}
          \end{minipage}\hfill
          \begin{minipage}[t]{0.49\linewidth}
            \lstinputlisting[style=pyhex-appendix,
              firstline=\number\hex@start,
              firstnumber=\number\hex@start]{#2}
          \end{minipage}
        \end{minipage}%
      }
    \endgroup
    \caption{#4}
    \label{#5}
  \end{figure}
}

\title{Agents of Change: Self-Evolving LLM Agents for Strategic Planning}

\author{Nikolas Belle$^{*}$, Dakota Barnes$^{*}$, Alfonso Amayuelas,\\
\textbf{Ivan Bercovich, Xin Eric Wang \& William Wang} \\
University of California, Santa Barbara \\
\texttt{\{nbelle, dakotabarnes, amayuelas\}@ucsb.edu},\\ 
\texttt{ibercovich@gmail.com}, \texttt{ericxwang@ucsb.edu},
\texttt{william@cs.ucsb.edu}
}

\iclrfinalcopy % Uncomment for camera-ready version, but NOT for submission.

\newif\ifshowfigs
\showfigstrue      % <— change to \showfigsfalse to hide all figures
\begin{document}
\maketitle

\begin{abstract}
We address the long-horizon gap in large language model (LLM) agents by enabling them to sustain coherent strategies in adversarial, stochastic environments. \textit{Settlers of Catan} provides a challenging benchmark: success depends on balancing short- and long-term goals amid randomness, trading, expansion, and blocking. Prompt-centric LLM agents (e.g., ReAct, Reflexion) must re-interpret large, evolving game states each turn, quickly saturating context windows and losing strategic consistency. We propose \textit{HexMachina}, a continual learning multi-agent system that separates environment \textit{discovery} (inducing an adapter layer without documentation) from strategy \textit{improvement} (evolving a compiled player through code refinement and simulation). This design preserves executable artifacts, allowing the LLM to focus on high-level strategy rather than per-turn reasoning. In controlled \textit{Catanatron} experiments, HexMachina learns from scratch and evolves players that outperform the strongest human-crafted baseline (AlphaBeta), achieving a 54\% win rate and surpassing prompt-driven and no-discovery baselines. Ablations confirm that isolating pure strategy learning improves performance. Overall, artifact-centric continual learning transforms LLMs from brittle stepwise deciders into stable strategy designers, advancing long-horizon autonomy.
\begingroup
\renewcommand\thefootnote{*}\footnotetext{Equal contribution}
\endgroup
\vspace{-10pt}
\end{abstract}

\section{Introduction}
Prompt-centric LLM agents and multi-agent systems are powerful but struggle on long-horizon tasks: as episodes unfold, prompts saturate with state summaries and improvised "memory," forcing the model to re-interpret its environment at every step (\cite{aghzal2025largelanguagemodelsgood, nayak2025llamarlonghorizonplanningmultiagent, chen2024can}). Achieving autonomous task following requires agents that do not need to relearn their interface at each inference step (\cite{bubeck2023sparksartificialgeneralintelligence, park2023generativeagentsinteractivesimulacra}). This has led to continual learning designs where LLMs embed feedback loops, revise prompts, and even generate tools or code to improve over time (\cite{zelikman2022starbootstrappingreasoningreasoning}). Allowing agents to gather and preserve artifacts (e.g., reusable functions or typed helpers) shifts heavy context parsing to deterministic code, freeing the model to focus on strategy rather than re-describing the world.

Despite this progress, few benchmarks test whether agents can refine coherent strategies over long horizons. Most existing domains emphasize short tasks or broad skill discovery, offering little insight into whether an agent can sustain and improve a single competitive policy. Yet this ability is crucial: real-world applications require agents not just to explore but to commit to strategies that hold over many steps amid uncertainty and competition. A benchmark demanding persistent strategy refinement against a strong adversary is therefore vital to assess whether lifelong agents truly close the long-horizon gap.

Settlers of Catan is an ideal stress test: each turn presents a large, evolving state and action space; success depends on balancing short- and long-term rewards under stochastic resource production, trading, expansion, and adversarial play. Using the open-source Catanatron framework (\cite{Collazo2025}) gives us a controlled interface to observe how a lifelong architecture impacts performance in a domain that reliably exposes limits in long-horizon reasoning.

We first demonstrate that traditional per-turn LLM agents (e.g., ReAct/Reflexion-style) perform poorly against a strong human-crafted bot. Asking the model to parse the full game state and independently choose every action while attempting to "hold" a global plan proves unreliable and inconsistent (Table~\ref{tab:player_comparison}). To address this, we separate the act of \textit{thinking} from the act of \textit{playing}, drawing inspiration from the AutoGPT framework (\cite{yang2023autogptonlinedecisionmaking}) to define distinct agent roles: Orchestrator, Analyst, Strategist, Researcher, and Coder. In this configuration, the system hypothesizes a strategy, translates it into a player implementation, reviews the API to ensure correctness, and then evaluates and improves through repeated play. While this Voyager-style (\cite{wang2023voyageropenendedembodiedagent}) continual learner shows progress, it tends to converge on shallow heuristics that fail to capture the depth of strategic play required in Catan (Appendix~A.2). Motivated by this limitation, we introduce a clean separation between the discovery of executable API artifacts and the refinement of strategies built on top of them. With this split, our system, \textit{HexMachina}, evolves players that consistently execute intelligent, long-horizon strategies,outperforming traditional LLM agents, common continual learning architectures, and even the AlphaBeta baseline.

\textbf{Main Contributions}. We highlight the following key contributions from our work:

\begin{itemize}
    \item \textbf{HexMachina: Self-Evolving LLM Agent Framework.} An autonomous system that learns an unknown environment without formal documentation, preserves key code/knowledge as artifacts, and improves its strategy via a closed-loop process that generates and executes code with no human intervention.
    \item \textbf{A strong benchmark setting for continual LLM-agent learning: \textit{Settlers of Catan}.} An environment that both requires long-horizon strategy and distracts naive agents with a large, changing state/action space and delayed rewards.
    \item \textbf{Lifelong agents beat traditional LLM agents on Catan.} HexMachina outperforms prompt-driven baselines and rivals the best human-engineered Catanatron bot (AlphaBeta) by letting the LLM design strategy while compiled code executes it consistently.
    \item \textbf{Empirical importance of separating discovery and improvement.} We show that decoupling environment-artifact discovery from strategy refinement materially improves strategy quality and game performance.
\end{itemize}

\ifshowfigs
\begin{figure}
    \centering
    \includegraphics[width=.98\textwidth]{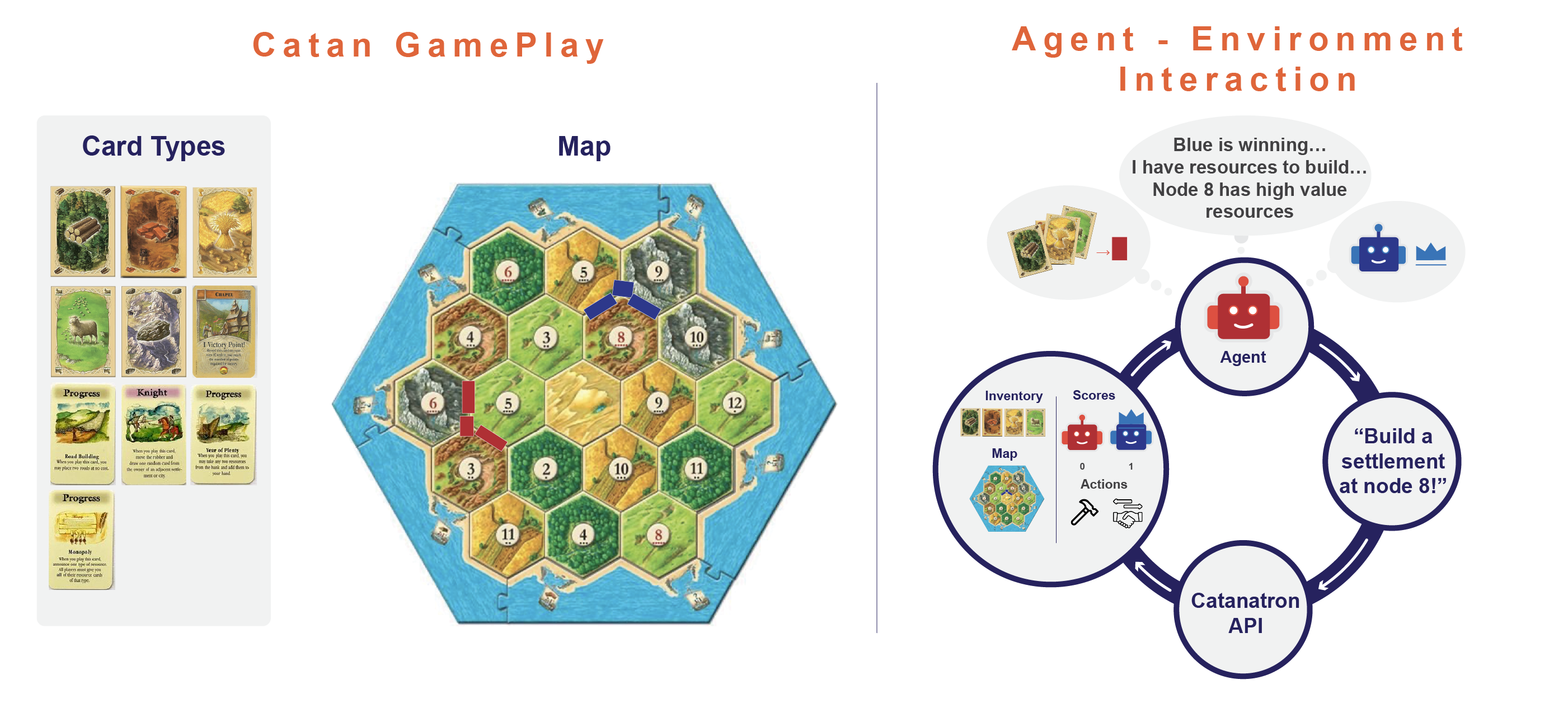}
    \caption{ \textbf{Overview of Catan gameplay and LLM-agent interaction}.
\textbf{Left}: \textit{Settlers of Catan} -- Players take turns to gather, trade, and spend resources to build on a modular board in a stochastic, partially observable strategy game. The objective is to reach 10 victory points by constructing settlements, roads, and cities \cite{catanfusion_blog, catancollector_guide}. \textbf{Right}: Our LLM-based framework  interacts with the Catanatron API, leveraging game state information and strategic reasoning to decide actions. Through repeated play and self-modification, agents evolve more coherent long-term strategies (\cite{flaticon_trade, flaticon_conversation, flaticon_thoughtbubble, flaticon_robot, flaticon_build}).}
    \label{fig:catan-fig}
    \vspace{-10pt}
\end{figure}
\fi

\section{Related Works}

\paragraph{Game-Playing AI and Strategy Games}  
Games have long served as benchmarks for AI research (\cite{gallotta2024large, costarelli2024gamebench, nasir2024gametraversalbenchmark}). While significant progress has been made in perfect-information games like Chess and Go (\cite{schultz2024mastering, silver2016mastering}), strategic board games such as \textit{Settlers of Catan}, \textit{Diplomacy} (\cite{meta2022human}) or \textit{Civilization} (\cite{qi2024civrealm}) introduce elements of expanding action spaces, partial observability, and multi-agent interaction, posing unique challenges to an AI system (\cite{szita2009monte}). Previous works approached Catan using a specialized neural network architecture to handle its mixed data types, enabling an RL agent to outperform traditional rule-based bots (\cite{gendre2020playingcatancrossdimensionalneural}). In contrast, our approach leverages LLMs' natural language understanding to navigate Catan's complexities, focusing on autonomous game-play discovery and strategy refinement without relying on extensive training data.

\paragraph{LLM Agents and Long-Horizon Planning}
LLMs reason well locally but falter at multi-step autonomy: studies report low success on end-to-end plan generation, with models performing better as advisors to external planners \cite{valmeekam2023planningabilitieslargelanguage}. Benchmarks like TravelPlanner confirm poor pass rates even with tools and staged prompting, revealing brittleness under constraint-heavy, multi-objective tasks \cite{xie2024travelplannerbenchmarkrealworldplanning, zheng2025lifelonglearninglargelanguage, nayak2025llamarlonghorizonplanningmultiagent, cui2025empoweringllmsparameterizedskills}. Prompt-centric agents (ReAct, Reflexion) still act per-turn from ever-growing text context, and multi-agent scaffolds (CAMEL, AutoGen) coordinate via dialogue \cite{yao2023react, shinn2023reflexionlanguageagentsverbal, wei2023chainofthoughtpromptingelicitsreasoning, xi2025rise, li2023camelcommunicativeagentsmind, wu2023autogenenablingnextgenllm}; yet in long-horizon, adversarial domains they repeatedly re-parse large states and lack a persistent executable substrate to enforce strategy across an episode, leaving the planning gap largely intact.

\paragraph{Self-Improvement and Continual Learning Agents}  
Inference-time self-improvement spans verbal reflection (Reflexion), evolutionary prompt search (PromptBreeder, PromptAgent), and code-writing agents that iteratively refine programs (\cite{shinn2023reflexionlanguageagentsverbal, fernando2023promptbreederselfreferentialselfimprovementprompt, wang2023promptagent}). Surveys systematize these inference-time strategies and the broader landscape of LLM agents (\cite{song2024mind, dong2024survey}). Eureka (\cite{ma2024eurekahumanlevelrewarddesign}) explores program and reward evolution, demonstrating how automated search over reinforcement learning environments can uncover novel control strategies. AlphaEvolve (\cite{novikov2025alphaevolve}) presents an evolutionary coding agent to tackle open scientific problems and algorithm improvement. Embodied lifelong systems like Voyager show that storing executable skills (a skill library) improves persistence and reuse across episodes, but emphasize breadth (discovering many primitives) rather than depth (refining a single competitive policy).

Building on Voyager, Eureka, and AlphaEvolve, which respectively advance skill discovery, reward/program evolution, and automated code improvement, we shift focus to a different question: can a lifelong LLM system, operating without documentation, induce a compact adapter to an unknown environment and persist executable artifacts in order to evolve a single competitive policy that outperforms traditional LLM agents in adversarial play?

% TODO Check to see if if alphaEvolve/funsearch should one or the other

% Sections Brainstorm: MAS, 
\ifshowfigs
\begin{table}[t]
\caption{\textbf{Focus comparison.} \checkmark=yes, \(\sim \)=partial, $\times$=no. Policy evolution (broad: direct or via reward/program/skill search); Artifacts (persisted executable code/skills); Induction (doc-free adapter induction; Voyager \(\sim \) with provided control primitives); Adversarial strategy-based (head-to-head vs strong fixed opponent; \checkmark only for HexMachina).}
\label{tab:systems_capabilities}
\begin{center}
\begin{tabular}{llcccc}
\multicolumn{1}{c}{\bf System}  &\multicolumn{1}{c}{\bf Environment} &\multicolumn{1}{c}{\bf Induction} &\multicolumn{1}{c}{\bf Artifacts} &\multicolumn{1}{c}{\bf Adversary} &\multicolumn{1}{c}{\bf Evolution}
\\ \hline \\
Voyager           & Minecraft  & \(\sim \) & \checkmark & $\times$ & \checkmark \\
AlphaEvolve & Code & $\times$ & \checkmark & $\times$ & \checkmark \\
Eureka            & Isaac Gym  & $\times$ & \checkmark & $\times$ & \checkmark \\
\bf HexMachina & Catanatron & \checkmark & \checkmark & \checkmark & \checkmark \\
\end{tabular}
\end{center}
\end{table}
\fi

\section{Background}

\paragraph{Settlers of Catan as a Strategic Benchmark}

\textit{Settlers of Catan} is a 3-4 player board game where players collect and trade resources to build settlements and roads, racing to earn 10 victory points on a modular island map. The game emphasizes resource management, planning, and negotiation, with mechanics like the robber (which blocks resources) adding tactical depth. Catan is known for its balance of luck and skill. \textbf{Victory} goes to the first player to reach \textbf{10 points}, earned by building and upgrading settlements into cities, buying development cards, and achieving goals like the longest road or largest army. Each settlement is worth 1 point, each city 2, and some development cards grant hidden points or knight bonuses. \textbf{Every turn} starts with a dice roll that produces resources for players with adjacent settlements. The active player may then trade and build. If a 7 is rolled, the robber is activated, blocking a tile and stealing a resource. Players must plan expansions, balance upgrades, and trade strategically to manage luck. This need for adaptation and foresight makes Catan a strong benchmark for evaluating strategic reasoning in agents.

\paragraph{The Catanatron Framework}

We use the open-source Python-based simulator \textbf{Catanatron} as our evaluation environment. Designed for automated gameplay of \textit{Settlers of Catan}, Catanatron offers a programmatic interface for integrating custom agents and supports rapid simulation at scale. It faithfully implements the game's rules and dynamics, capturing key strategic elements such as resource management, trade negotiation via structured proposals, and randomness introduced by dice rolls. Each game consists of players competing to reach ten victory points, with players interacting through well-defined game states that include current resources, board positions, available actions, and observable opponent statuses. Games typically span 40 to 100 turns, allowing for extended observation of agents' long-term planning capabilities. We benchmark our LLM-driven agents against \textbf{AlphaBeta}, the best-performing heuristic agent provided through the API which uses a depth-2 alpha-beta pruning algorithm with heuristic evaluation to select actions.

\paragraph{Alpha-Beta Benchmark}  

Our primary baseline is Catanatron's AlphaBeta agent: an alpha-beta minimax over stochastic outcomes that computes the expected value of successor states via chance expansion and a fast heuristic value function. Concretely, it uses a depth-2 search (default), a 20 s decision cap, and an optional action-space pruning mode (e.g., robber and maritime-trade pruning heuristics). At leaves, it applies a parameterized value function, and it short-circuits when only one legal action exists. We adopt the author defaults unless otherwise noted, fixing depth = 2 for all reported comparisons. This baseline is both strong and extremely fast, enabling thousands of head-to-head evaluations needed by our continual-learning setup.

\section{HexMachina}
% High level description of our system. Key Points:\\
% - We press run on the system, and no human intervention until it finishes with evolved player. (fully autonomous)
% - Run through an example workflow briefly. Start with Template, Run Game, Analyzer, Meta Chooses, Researcher, Meta Chooses, Coder, Run Game, Analyzer...\\
% - Use 20 steps per evolution run

HexMachina is an autonomous self-evolving multi-agent system that crafts a powerful Catanatron player capable of rivaling the top human-crafted baselines. We utilized Langchain for the model agnostic services, and Langraph for the state machine. Once launched, HexMachina begins by running a \textit{discovery phase}, were it gathers information about the Catanatron API to evolve an \textbf{adapters} file. After completion, it enters an \textit{improvement phase} where it begins evolving a \textbf{player} file. Each evolution consists of agent collaboration until the Coder writes improvements in the form of testable code. Each phase is limited to 20 evolutions, counted by each time the Coder is called.

\ifshowfigs
\begin{figure}[h]
\centering
\includegraphics[width=1.0\linewidth]{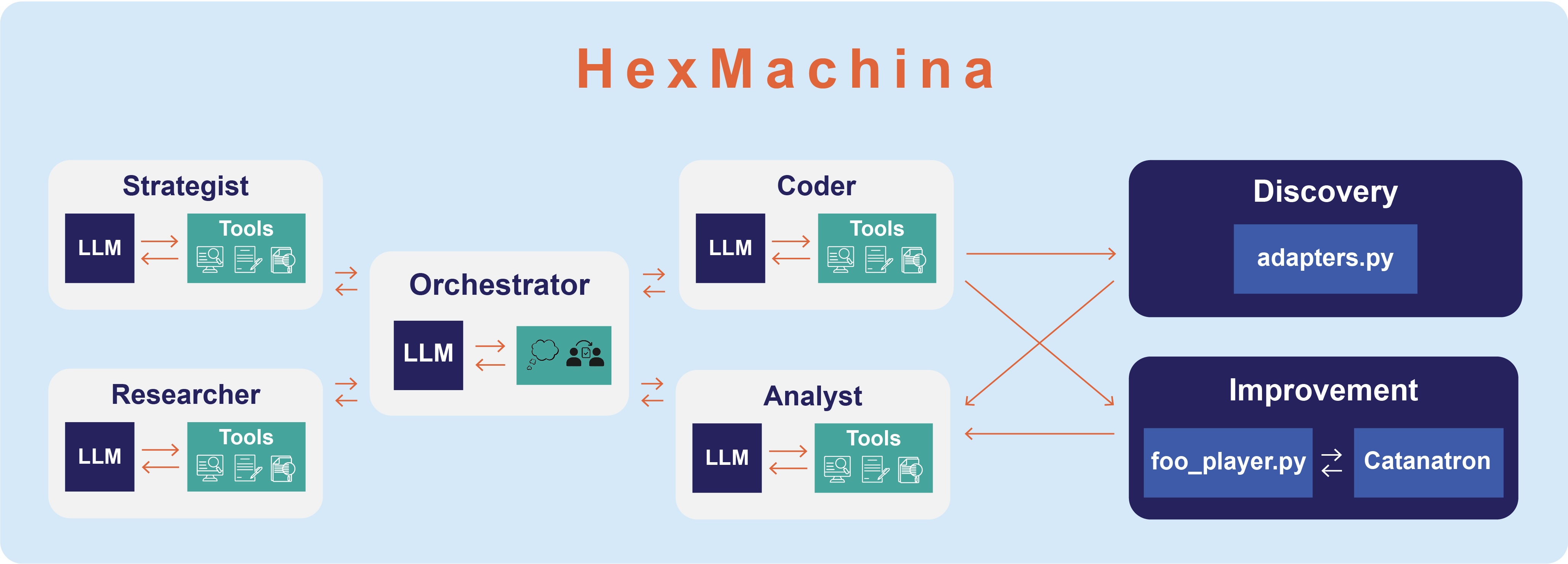}
\caption{\textbf{HexMachina Architecture.} During the discovery phase, the Orchestrator coordinates agents to induce executable functions into the \texttt{Adapter} file, stabilizing access to the environment. In the improvement phase, we found it most effective to rely on a streamlined loop of \textit{Analyst}, \textit{Coder}, and \textit{Orchestrator}, avoiding dilution from additional roles. Here, the Analyst diagnoses performance, the Coder translates revisions into executable code, and the Orchestrator manages iteration. This separation enables the system to refine \texttt{FooPlayer} into a consistent long-horizon strategy.}

\label{fig:hexmachina_diagram}
\end{figure}
\fi

\subsection{Capabilities}
% Short paragraph What it can do: Capabilities that the system implements (descriptions of how and benefit, keys from comparison figure)
Listed below are the capabilities that enable HexMachina to employ continual learning effectively:

\paragraph{Player Generation}
HexMachina incrementally codes a complete Catanatron \textbf{player} module during the \textit{improvement phase}. The process begins with a minimal template that simply returns the first legal action, then evolves into increasingly sophisticated strategies as feedback accumulates. Importantly, the generated player is not just a script of next actions but an executable policy that can consistently carry out a long-term plan across an entire game. This design shifts the LLM's role from being a per-turn decider to being a strategy architect, with the Coder agent ensuring that every idea is grounded in syntactically valid and testable code.

\paragraph{Experimentation Engine}
At the core of HexMachina is a deterministic experimentation harness. This engine repeatedly pits evolving players against the strong AlphaBeta baseline under fixed seeds and identical settings, logging outcomes, intermediate states, and decision traces. By holding the environment constant, we can attribute changes in win rate or victory points directly to code evolution, rather than stochastic noise. This repeatable evaluation cycle transforms raw self-play into structured experimental evidence for policy improvement.

\paragraph{Evaluation}
Evaluation closes the loop: after each batch of games, HexMachina analyzes outcomes to identify which strategic choices were beneficial and which led to failure. This feedback is distilled into concise summaries that the Orchestrator uses to decide whether to preserve, modify, or discard a candidate player. In practice, we found that the most effective evaluation loop did not require every agent's perspective; instead, a streamlined pipeline of \textit{Analyst to Coder to Orchestrator} yielded clearer strategic signals. Additional recommendations from other roles often diluted the strategy, fragmenting the LLM's ability to commit to a coherent plan.

\paragraph{Strategy and Discovery}
HexMachina supports two complementary modes. In the \textit{discovery phase}, it induces executable artifacts such as an \texttt{adapters.py} file that stabilizes access to the Catanatron API without any human documentation. This ensures that later improvements build on a reliable, reusable interface. In the \textit{improvement phase}, the system searches over new tactics, revisits prior players, and integrates insights from past runs. Together, these phases ensure that learning is both grounded in the environment and continuously refined across evolutions.

\paragraph{Orchestration}
The orchestrator serves as the global planner, deciding when to analyze results, request new code, or revisit prior knowledge. Autonomy is enforced by a closed loop: the orchestrator makes high-level decisions based on game outcomes, artifacts, and agent communication, then delegates low-level tasks to the Analyzer and Coder. This separation prevents the system from stalling on details while still maintaining tight control over long-term strategy evolution.

\paragraph{Memory}
Finally, HexMachina maintains both \textit{game memory} and \textit{semantic memory} across evolutions. Game memory archives past players, their code, and evaluation artifacts, enabling direct comparisons and reuse of successful strategies. Semantic memory allows each agent to persist expertise relevant to its role, e.g., the Coder retaining knowledge of syntax patterns or the Analyst preserving diagnostic heuristics. This dual memory system underpins continual learning: instead of starting from scratch each evolution, the system accumulates strategic and technical knowledge that compounds over time.

% \subsection{Agents}
% % Short Paragraph about the agents in our MAS. Key Points:\\
% % - Mention how each agent is able to call tools (max 5?), before returning back to the main loop with the final findings/message. Their tool call history for a turn is not included in their total memory, just the final one.\\
% % - Each Agent should have information about their inputs, outputs, system role, tools available
% Each agent is a specialist that can use tools (<6 per turn) and then yields a single, compact message back to the main loop; only that final message is persisted to memory. Each agent has access to a \textit{Think Tool}, (from the Langchain deep research agent) that elicits chain-of-thought for the agents decision making and explainability. Inputs and outputs are standardized with Langchain, enabling interchangeable models and providers.
\subsection{Agents}
Each agent in HexMachina is a specialist that can call tools (up to 5 per turn) and then yield a single, compact message back to the main loop. Only this final message is persisted to memory, ensuring concise, role-specific contributions. Each agent has access to a \textit{Think Tool} (adapted from Langchain's deep research agent) to support internal reasoning and explainability. Inputs and outputs are standardized, enabling interchangeable models and providers.

Importantly, not all agents are equally useful in every phase. In the \textit{discovery phase}, the full set of agents contributes to inducing a stable \texttt{adapters.py} file from scratch. However, in the \textit{improvement phase}, we found that the most effective configuration is a streamlined loop of \textbf{Orchestrator, Analyst, and Coder}. Additional recommendations from the Strategist and Researcher often diluted coherence, so these roles are used only during discovery or when revisiting artifacts, not for direct strategy refinement.

% \paragraph{Meta.} \emph{Role:} global planner/orchestrator. \emph{Inputs:} last analysis, current artifacts, recent metrics. \emph{Outputs:} next action plan (research/strategy/code/eval) with tool budget. \emph{Tools:} run games, read logs, list/edit files, unit tests, repo search.

% \paragraph{Coder}
% Agent role in system.

% \paragraph{Analyzer}
% Agent role in system.

% \paragraph{Researcher}
% Agent role in system.

% \paragraph{Strategist}
% Agent role in system.

% Compact agent spec list item
\newcommand{\AgentSpec}[5]{%
  \item[\textbf{#1}:] #2\\
  \textit{Inputs:} #3\\
  \textit{Outputs:} #4\\
  \textit{Tools:} #5%
}
\begin{description}
  \AgentSpec{Orchestrator}
    {Global planner and orchestrator.}
    {Orchestrator Messages and summary of evolution}
    {Thoughts, system goal, next chosen agent, next agent objective}
    {None}

  \AgentSpec{Coder}
    {Turn strategies into compilable code.}
    {Objective from Orchestrator, adapter contents}
    {Executable code, summary of changes}
    {Write/edit file}

  \AgentSpec{Analyst}
    {Experimentation evaluator}
    {Objective from Orchestrator, summary of evolution, current player and Coder summary of changes, game artifacts, adapter contents}
    {Post-game diagnosis, specific analysis, adapter failure}
    {Read local file}

  \AgentSpec{Researcher}
    {Recover API/engine facts and domain tactics. Primarily active during discovery.}
    {Objective from Orchestrator, list of files, adapter contents}
    {Citations, code pointers, or concise notes with source references}
    {Read local file, web search}

  \AgentSpec{Strategist}
    {Propose concrete, testable plans. Primarily active during discovery.}
    {Orchestrator Objective, Evolution Summary, current player, adapter contents}
    {Strategy spec and evaluation}
    {Read local file, view older experiment, web search}

\end{description}

\section{Experiment Setup}
We evaluate HexMachina in the open-source \textit{Catanatron} environment under controlled 2-player, 
10-point Catan games. Each experiment consists of repeated head-to-head matches against the strongest built-in heuristic bot, \textit{AlphaBeta}. 
We measure both \textit{win rate} and \textit{final victory points} as indicators of strategic quality. 
Games are deterministic given a random seed, allowing us to reproduce results and separate genuine improvements from stochastic variance. 
Data was collected over 60 hours across two machines (MacBook Pro 2019, 16GB; MacBook M1 Max 2021, 32GB).  

\subsection{Baselines}
Our baselines capture a spectrum of reference points, from trivial random play to a strong, hand-engineered heuristic, allowing us to contextualize HexMachina's performance against both naive policies and established rule-based expertise.

\paragraph{Random.}
The simplest control agent chooses uniformly from the legal action space each turn. 
While strategically meaningless, this baseline sets a lower bound for performance and highlights how much structure even a minimal policy adds.  

\paragraph{LLM Player.}
We also evaluate a Reflexion-style agent (\cite{shinn2023reflexionlanguageagentsverbal}) that reformats the game state into text and queries Claude 3.7 once per turn with a high-level goal. 
This baseline reflects the "prompt-centric" paradigm: the LLM directly drives play without any compiled memory or artifact reuse. 
Due to inference cost (approx. 70 queries per game), we limited this evaluation to 20 games with a model we had free access too, but it provides a critical comparison to show how quickly context saturation and lack of persistence hinder long-horizon play.  

\paragraph{Basic Continual Learner (HexMachina w/o discovery).}
To isolate the value of separating discovery and improvement, we also test a single-phase continual learning setup equivalent to HexMachina without the discovery phase. 
Here the system attempts to learn both the environment interface and the strategy simultaneously. 
This resembles prior lifelong agents such as Voyager and Eureka (\cite{wang2023voyageropenendedembodiedagent,ma2024eurekahumanlevelrewarddesign}), which evolve strategies directly from raw interaction. 
As shown later in Appendix~A.2, these agents often converge on shallow heuristics (e.g., one-ply VP-only evaluators), highlighting the difficulty of strategic refinement without first stabilizing the interface.  

\paragraph{AlphaBeta.}
Finally, we include Catanatron's AlphaBeta agent, a depth-2 minimax with stochastic expansion and heuristic evaluation. 
This player is fast, strong, and widely used as a benchmark; in self-play it achieves a 51\% win rate by construction. 
It represents the ceiling for our experiments, providing a human-engineered reference against which HexMachina's evolved players can be meaningfully compared.  

\subsection{Models}
HexMachina is model-agnostic, but in practice we deploy different LLMs for different roles to balance strength and efficiency. 
We test three orchestrator backends, GPT-5-mini, Claude 3.7, and Mistral-large, to assess robustness across providers. 
Unless otherwise noted, GPT-5-mini is used for the Coder, which requires reliable code synthesis, while Mistral-large is assigned to support roles (Analyst, Strategist, and Researcher) to reduce cost and latency. 
This division reflects a general principle of our framework: leverage stronger models where precision is critical (e.g., code generation) and more efficient models where interpretive or diagnostic reasoning suffices.

\section{Results and Discussion}

\subsection{Continual Learning}
We first examine the impact of continual learning through evolution runs of 10 steps, with each step evaluating FooPlayer across 30 games. Figure~\ref{fig:hexmachina-lineplot} shows HexMachina steadily improving against AlphaBeta, eventually achieving parity and surpassing baseline players. A central design choice was the separation of \textit{discovery} (API induction and artifact stabilization) from \textit{improvement} (strategy evolution). Our experiments confirm that this separation is critical: systems without discovery struggled to stabilize player code, while those with discovery reliably produced executable players that improved across evolutions. 

Interestingly, we found that HexMachina performed better when the Strategist and Researcher agents were \textit{removed}, leaving only the Orchestrator, Analyst, and Coder. While the Strategist was intended to propose concrete plans, results suggest that LLMs often formulate effective strategies in a single shot, and passing these through multiple roles may dilute coherence. Thus, we report results using this streamlined configuration. This insight highlights a broader implication for continual learning: modular multi-agent systems are powerful, but not all roles contribute equally, and reducing mediation can strengthen strategic consistency.

Figure~\ref{fig:evolution_dialogue} provides a qualitative example of evolution in action. We observe HexMachina iteratively proposing, coding, and refining player strategies while preserving functional artifacts. This illustrates how artifact-centric continual learning transforms an LLM from a per-turn decision maker into a higher-level strategy designer with consistent policy execution.

\ifshowfigs
\begin{figure}[h]
\centering
\includegraphics[width=0.8\linewidth]{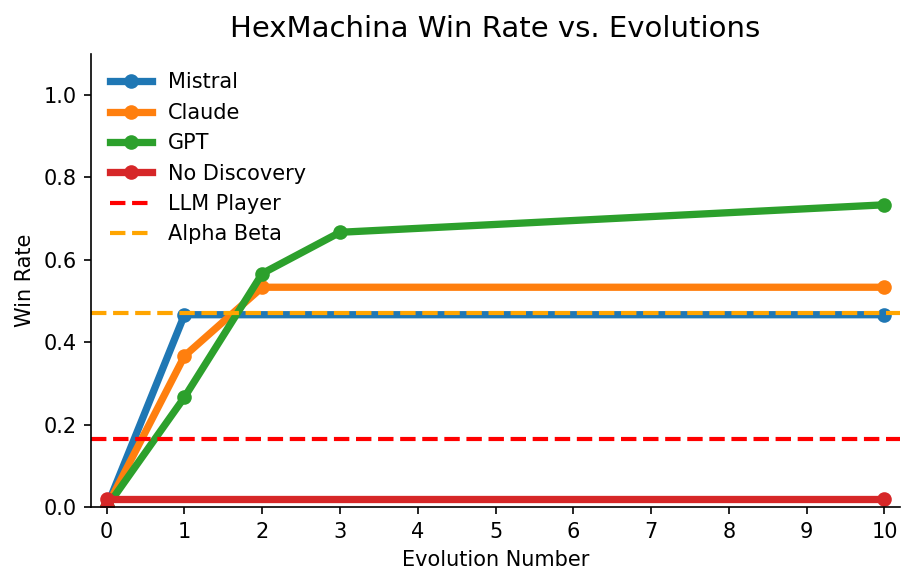}
\caption{HexMachina's Best-So-Far Performance Curve Outpacing Existing Players.}
\label{fig:hexmachina-lineplot}
\end{figure}
\fi

\ifshowfigs
\begin{figure}[h]
\centering
\includegraphics[width=1.0\linewidth]{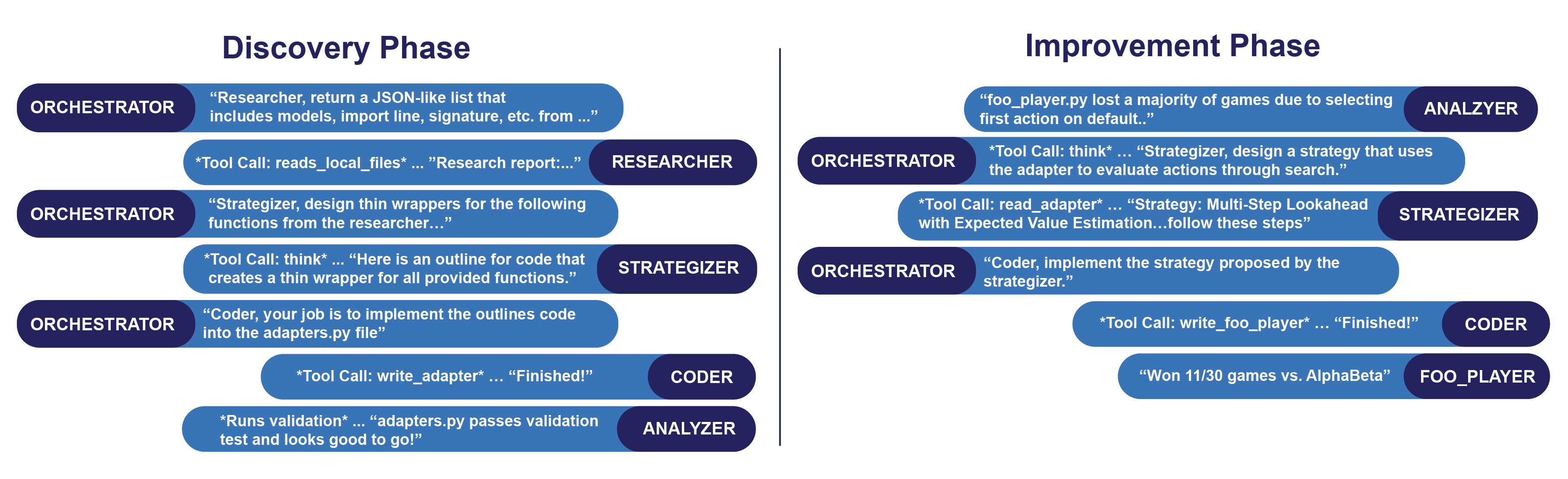}
\caption{Evolution Messages Example Dialogue}
\label{fig:evolution_dialogue}
\end{figure}
\fi

\subsection{Player Comparison}
To stress test the best-evolved players, we ran each configuration 10 times with 100 games per run. Results are summarized in Table~\ref{tab:player_comparison}. HexMachina's best model (GPT-5-mini) reached a 54.1\% win rate and 8.2 $\pm$ 0.1 victory points, matching or slightly exceeding AlphaBeta's 51.0\% win rate and 7.8 $\pm$ 0.2 points. By contrast, the no-discovery baseline plateaued at much lower win rates, producing players that often failed to generalize beyond static heuristics.

A representative no-discovery agent (Appendix~A.2) highlights why this baseline performs poorly. It carries out only a 1-ply lookahead, scoring states almost entirely on current victory points with trivial tie-breakers such as settlements, cities, or roads. With rollouts disabled and no modeling of stochastic production or opponent actions, it assigns identical scores to materially different choices, leading to random tie-breaking, poor settlement placement, and ineffective robber usage. These flaws explain its consistently weak performance in Table~\ref{tab:player_comparison}.  

By contrast, the best evolved \texttt{FooPlayer} (Appendix~A.1) demonstrates the benefits of HexMachina's discovery-improvement split. This agent combines phase-aware priorities (early expansion, mid-game balance, late-game upgrades), explicit heuristics for production diversity and robber disruption, and shallow rollouts that anticipate near-term outcomes. These capabilities yield stronger growth, better-timed upgrades, and consistent disruptive pressure on the opponent. The qualitative differences map directly onto the quantitative results in Table~\ref{tab:player_comparison}, underscoring that discovery is critical for stabilizing adapters and enabling the emergence of richer strategies.

\ifshowfigs
\begin{table}[h]
\caption{Win Rate and Victory Points for HexMachina compared to Baselines}
\label{tab:player_comparison}
\centering
\begin{tabular}{llll}
\multicolumn{1}{c}{\bf Player} & \multicolumn{1}{c}{\bf Model} & \multicolumn{1}{c}{\bf Win Rate} & \multicolumn{1}{c}{\bf Victory Points} \\
\addlinespace[2pt]
\midrule[0.2pt] % very thin separator; adjust thickness as desired
\addlinespace[2pt]
\multirow{3}{*}{HexMachina}
  & \textbf{GPT}          & \textbf{54.1\% [51\%, 57\%]} & \textbf{8.2 $\pm$ 0.1} \\
  & Mistral      & 49.2\% [46\%, 52\%] & 7.8 $\pm$ 0.2 \\
  & Claude       & 38.4\% [35\%, 41\%] & 7.2 $\pm$ 0.2 \\
\addlinespace[2pt]
\midrule[0.2pt] % very thin separator; adjust thickness as desired
\addlinespace[2pt]
\multirow{1}{*}{LLM Player}
  & Claude      & 16.4\% [3\%, 30\%]  & 5.2 $\pm$ 1.2 \\
\addlinespace[2pt]
\midrule[0.2pt] % very thin separator; adjust thickness as desired
\addlinespace[2pt]
\multirow{1}{*}{Alpha-Beta}
  & X      & 51.0\% [48\%, 54\%] & 7.8 $\pm$ 0.2 \\
\addlinespace[2pt]
\midrule[0.2pt] % very thin separator; adjust thickness as desired
\addlinespace[2pt]
\multirow{1}{*}{Random}
  & X      & 0.2\% [0\%, 0\%] & 2.4 $\pm$ 0.0 \\
\end{tabular}
\end{table}
\fi

% \subsection{Ablations}
% Section about DESIGN CHOICES we made, and what happens if we remove them\\
% Experiment of running each ablation 5 times, 20 steps, 100 games each for error bars\\
% Utilize HexMachina Avg and Best for the comparison.

% Remove each agent from the improvement phase assuming a stable adapters.py file to see the impact they each have on policy evolution. Researcher and strategist we can remove as an option for Orchestrator\\
% Analyst have to remove from react graph after game run\\
% Include discussion on based on table results and findings
% explain how removing strategist performed better than when it was present in improvement. explain why this might've been.

\subsection{Ablations}
To isolate the importance of individual design choices, we conducted ablation studies with three independent runs of 10 steps, each tested on 30 games. Results are shown in Table~\ref{ablations}. Interestingly, removing the Strategist and Researcher improved performance relative to the full system, reaffirming our earlier finding that direct orchestration leads to clearer strategy translation. Removing the Analyst heavily impacted success as the agent is required to diagnose issues. There would often be situations where the system failed to recognize when functions were being mis-referenced from \texttt{adapters.py} without the Analyst bringing it into a failure loop.
\ifshowfigs
\begin{table}[h]
\caption{Multi-Agent Architecture Ablations for HexMachina Policy Evolution}
\label{ablations}
\begin{center}
\begin{tabular}{lll}
\multicolumn{1}{c}{\bf Ablation}  &\multicolumn{1}{c}{\bf Win Rate} &\multicolumn{1}{c}{\bf Victory Points} \\
% \\ \hline \\
% No Strategist          &$50\% \pm 1\%$     &$8.0\% \pm 1\%$\\
% No Researcher          &$50\% \pm 1\%$     &$8.0\% \pm 1\%$\\
% No Analyst             &$50\% \pm 1\%$     &$8.0\% \pm 1\%$\\
% %No thinking tool?
% \hline
% \textbf{HexMachina}          & \textbf{50\% $\pm$ 1\%}    & \textbf{8.0\% $\pm$ 1\%}\\
\addlinespace[2pt]
\midrule[0.2pt] % very thin separator; adjust thickness as desired
\addlinespace[2pt]
All Agents             &49.7\%   &8.0\\
No Analyst             &0.0\%    &2.1\\
\addlinespace[2pt]
\midrule[0.2pt] % very thin separator; adjust thickness as desired
\textbf{No Strategist + Researcher}          &\textbf{54.1}\%   &\textbf{8.2}\\
\addlinespace[2pt]
\end{tabular}
\end{center}
\end{table}
\fi
Overall, these findings back our contribution statements: (1) HexMachina evolves executable strategies that rival top human-crafted baselines; (2) artifact preservation and doc-free discovery are essential to this success; (3) LLMs are best deployed at the level of strategy design, not per-turn play; and (4) multi-agent modularity is powerful, but optimal performance may emerge from leaner configurations that avoid unnecessary role handoffs.

% \section{Conclusion}
% One paragraph about what hindered us to not do better. Key Points:\\
% - Only use VP and Win Rate to determine which player is “better”\\
% - Sometimes the model can hallucinate\\
% - Expensive due to cost\\
% - Performance tied to model quality\\
% - Final sentence on with all of our limitations, what we were still able to achieve
% One paragraph on how this paper can be taken to the next level. Key Points:\\
% - Other people should try and create a better MAS on this benchmark, or a better human crafted player\\
% - Bring out MAS to new continual learning benchmarks\\
% - Extend number of steps (currently 20) that the model can evolve through. Needs better memory and player management to do this.

\section{Conclusion}

Despite strong results, several limitations hindered performance from improving further. First, we evaluated players solely with win rate and final victory points, coarse metrics that sometimes mask subtler strengths and weaknesses. Second, the LLM occasionally hallucinated code or heuristics, requiring additional filtering. Third, the system was expensive to run due to inference costs, restricting the number of trials. Finally, performance remained closely tied to the quality of the underlying model, with more capable backends producing stronger players. Even with these constraints, HexMachina was able to autonomously induce an API, evolve a robust player, and achieve parity with AlphaBeta, the strongest human-crafted bot.

Looking forward, we see several avenues for advancement. Other researchers could attempt to design a more powerful multi-agent system on this benchmark or build a stronger hand-crafted player for comparison. More broadly, HexMachina should be tested on continual learning benchmarks beyond Catan to validate generality. Finally, the current 20-step evolution limit could be extended with improved memory and player management, enabling longer training horizons and more sophisticated strategies. Together, these extensions would push LLM agents closer to reliable long-horizon autonomy.

% NOTE: ETHICAL STATEMENT and REPRODUCIBILITY STATEMENT are not in page count
\section{Ethical Statement}
Our system executes code in a closed loop with strict safeguards: generated programs run only within a controlled evaluation harness, preventing arbitrary system access. All experiments were logged with fixed random seeds and configuration files, ensuring transparency and reproducibility. While we present HexMachina as an autonomous agent, we avoid anthropomorphizing. Our system is a code-evolving tool, not a sentient entity.

\section{Reproducibility Statement}
We release all code, experiment harnesses, and configuration files alongside this submission. To reproduce our results, clone the repository, install dependencies, and follow the step-by-step README instructions. Running experiments requires API keys for the tested LLMs; once provided, the system can be executed exactly as described to replicate all tables and figures.

\bibliography{iclr2026_conference}
\bibliographystyle{iclr2026_conference}

\newpage
\appendix
\section{Appendix}

\subsection{HexMachina's Best Strategy}
\begin{python}
import random
from typing import Iterable, List, Optional, Any, Tuple

# MUST use the adapters surface to interact with the game environment
from .adapters import (
    Game,
    Player,
    Color,
    copy_game,
    execute_deterministic,
    execute_spectrum,
    expand_spectrum,
    list_prunned_actions,
    prune_robber_actions,
    base_fn,
    value_production,
    get_value_fn,
)

class FooPlayer(Player):
    """A Foo player with game-phase aware decisioning, improved sampling,
    short rollouts, and richer heuristics.

    This implementation is defensive: it uses only the adapters surface and
    contains many fallbacks when attributes or adapter helpers are missing.

    Key features:
    - Game-phase detection (early/mid/late) to bias settlement/road vs city/dev-card
    - Settlement & road potential heuristics to encourage early expansion
    - Robber/knight evaluation to value disruption and steals
    - Must-include guarantees for critical action types (settlement/road/robber/dev)
    - Rollout policy biased by phase and includes a light opponent-response

    NOTE: Many game model attribute names vary across environments. This code
    attempts multiple common attribute names and falls back to string-based
    heuristics when necessary. If the next run raises AttributeError for an
    adapters function or a specific attribute, provide the traceback so it can
    be patched to the concrete environment.
    """

    # Tunable constants (exposed to edit for experimentation)
    MAX_SIMULATIONS = 24
    PREFILTER_TOP_K = 8
    ROLLOUT_DEPTH = 2
    SIMULATION_BUDGET = 60
    DEBUG = False

    # Phase thresholds (used by get_game_phase)
    EARLY_TURN_THRESHOLD = 20
    MID_TURN_THRESHOLD = 45

    # Phase multipliers matrix (explicit)
    MULTS = {
        "EARLY": {"settlement": 2.0, "road": 1.8, "city": 0.8, "dev": 1.2},
        "MID": {"settlement": 1.0, "road": 1.0, "city": 1.25, "dev": 1.0},
        "LATE": {"settlement": 0.8, "road": 0.9, "city": 1.5, "dev": 1.0},
    }

    # Must-include action tokens (robust, lowercase matching)
    MUST_INCLUDE_TOKENS = {
        "build_city",
        "build_settlement",
        "build_sett",
        "build_road",
        "buy_dev",
        "buy_dev_card",
        "buycard",
        "play_knight",
        "knight",
        "move_robber",
        "move_robber_action",
        "robber",
        "trade",
        "offer_trade",
    }

    # Robber scoring base (increased)
    ROBBER_BASE_SCORE = 80.0
    ROBBER_BASE_SCORE_HIGH = 80.0

    # Settlement target in early game
    TARGET_SETTLEMENTS_EARLY = 3

    # Epsilon-greedy randomness to avoid predictability
    EPSILON_GREEDY = 0.04

    # Rollout bonuses for the very first rollout step
    ROLLOUT_SETTLEMENT_BONUS = 1.7
    ROLLOUT_ROAD_BONUS = 1.4

    # Tie tolerance
    TOLERANCE = 1e-6

    # Development card deck & EV constants
    DEV_DECK = {"knight": 14, "vp": 5, "road_building": 2, "year_of_plenty": 2, "monopoly": 2}
    DEV_TOTAL = sum(DEV_DECK.values())
    EV_KNIGHT = 0.15
    EV_VP = 1.0
    EV_ROAD_BUILDING = 0.25
    EV_YOP = 0.2
    EV_MONOPOLY = 0.3
    DEV_EV_SCALE = 60.0
    DEV_EV_THRESHOLD = 0.25

    # Knight bonuses
    KNIGHT_LARGEST_ARMY_BONUS = 50.0
    KNIGHT_BASE = 25.0
    KNIGHT_MIN_SCORE = 35.0

    # City/road/robber tuning (from latest analyzer guidance)
    CITY_URGENCY_BONUS = 85.0
    CITY_AFFORD_STRICT_ORE = 3
    CITY_AFFORD_STRICT_WHEAT = 2
    CITY_AFFORD_SOON_ORE = 2
    CITY_AFFORD_SOON_WHEAT = 1
    ROLLOUT_CITY_BONUS = 1.8
    ROAD_SCORE_BOOST = 9.0
    PROD_LOSS_IMPORTANCE = 70.0
    HIGH_VALUE_RESOURCE_SET = {"ore","wheat","metal","grain"}
    CITY_TIE_EPS = 0.02

    # Forcing behavior flags and diagnostic counters
    PREFILTER_FORCE_CITY_IF = True
    CITY_FORCE_AFFORD_STRICT = True
    DEBUG_COUNTS = False

    def __init__(self, name: Optional[str] = None):
        super().__init__(Color.BLUE, name)
        # Try to cache a base value function from adapters
        try:
            self._value_fn = base_fn()
            self.debug_print("FooPlayer: Using adapters.base_fn() for evaluation")
        except Exception as e:
            self._value_fn = None
            self.debug_print("FooPlayer: adapters.base_fn() not available, will use heuristic. Error:", e)

        # Diagnostic counters (quiet unless DEBUG)
        self._diag_forced_settlement = 0
        self._diag_forced_road = 0
        self._diag_city_urgency_count = 0
        self._diag_settle_urgency_count = 0

        # New counters for tuning
        self.COUNTER_FORCED_CITY = 0
        self.COUNTER_DEV_BUY_FORCED = 0
        self.COUNTER_BUY_DEV_ACTUALLY = 0
        self.COUNTER_BUILD_CITY_ACTUALLY = 0
        self.COUNTER_ROBBER_ACTUALLY = 0

    # ------------------- Debug helper -------------------
    def debug_print(self, *args: Any) -> None:
        if self.DEBUG:
            print(*args)

    # ------------------- Utility helpers -------------------
    def _get_player_color(self) -> Color:
        """Return this player's color. Try common attribute names."""
        if hasattr(self, "color"):
            return getattr(self, "color")
        if hasattr(self, "_color"):
            return getattr(self, "_color")
        return Color.BLUE

    def _safe_action_name(self, action: Any) -> str:
        """Produce a lowercase string name for the action for robust matching."""
        try:
            at = getattr(action, "action_type", None)
            if at is None:
                at = getattr(action, "type", None)
            if at is not None:
                try:
                    return str(at.name).lower()
                except Exception:
                    return str(at).lower()
        except Exception:
            pass
        try:
            # Some Action objects have a .name or .action_name
            name = getattr(action, "name", None) or getattr(action, "action_name", None)
            if name is not None:
                return str(name).lower()
        except Exception:
            pass
        try:
            return str(action).lower()
        except Exception:
            return ""

    # ------------------- Phase detection -------------------
    def get_game_phase(self, game: Game, color: Optional[Color] = None) -> str:
        """Return 'EARLY', 'MID', or 'LATE' based on turn counters or VP thresholds.

        Order of checks:
        1) turn/tick counters if available (preferred)
        2) max VP among players
        3) fallback to conservative MID
        """
        try:
            state = getattr(game, "state", game)
            turn_count = (
                getattr(state, "turn", None)
                or getattr(state, "tick", None)
                or getattr(state, "turn_count", None)
                or getattr(state, "tick_count", None)
            )
            if isinstance(turn_count, (int, float)):
                tc = int(turn_count)
                if tc < self.EARLY_TURN_THRESHOLD:
                    return "EARLY"
                if tc < self.MID_TURN_THRESHOLD:
                    return "MID"
                return "LATE"
        except Exception:
            pass

        # Fallback: use maximum VP among players
        try:
            state = getattr(game, "state", game)
            players = getattr(state, "players", None) or getattr(game, "players", None) or []
            max_vp = 0
            if isinstance(players, dict):
                for p in players.values():
                    vp = getattr(p, "victory_points", None) or getattr(p, "vp", None) or 0
                    try:
                        vp = int(vp)
                    except Exception:
                        vp = 0
                    max_vp = max(max_vp, vp)
            else:
                for p in players:
                    vp = getattr(p, "victory_points", None) or getattr(p, "vp", None) or 0
                    try:
                        vp = int(vp)
                    except Exception:
                        vp = 0
                    max_vp = max(max_vp, vp)
            if max_vp < 4:
                return "EARLY"
            if max_vp < 8:
                return "MID"
            return "LATE"
        except Exception:
            # Conservative fallback to MID
            return "MID"

    # ------------------- Heuristic / evaluation (phase-aware) -------------------
    def _heuristic_value(self, game: Game, color: Color) -> float:
        """Phase-aware heuristic including production potential and city-upgrade progress.

        Many attribute names are attempted to be robust across different game models.
        """
        # Die probabilities for numbers 2..12 ignoring 7
        die_prob = {2: 1 / 36, 3: 2 / 36, 4: 3 / 36, 5: 4 / 36, 6: 5 / 36, 8: 5 / 36, 9: 4 / 36, 10: 3 / 36, 11: 2 / 36, 12: 1 / 36}

        # Player lookup
        player_state = None
        try:
            state = getattr(game, "state", game)
            players = getattr(state, "players", None) or getattr(game, "players", None)
            if isinstance(players, dict):
                player_state = players.get(color) or players.get(str(color))
            elif isinstance(players, (list, tuple)):
                for p in players:
                    if getattr(p, "color", None) == color or getattr(p, "color", None) == str(color):
                        player_state = p
                        break
        except Exception:
            player_state = None

        def _safe_get(obj, *names, default=0):
            if obj is None:
                return default
            for name in names:
                try:
                    val = getattr(obj, name)
                    if val is not None:
                        return val
                except Exception:
                    try:
                        val = obj[name]
                        if val is not None:
                            return val
                    except Exception:
                        continue
            return default

        vp = _safe_get(player_state, "victory_points", "vp", default=0)
        settlements = _safe_get(player_state, "settlements", "settle_count", "settle_locations", default=0)
        if isinstance(settlements, (list, tuple)):
            settlements = len(settlements)
        cities = _safe_get(player_state, "cities", "city_count", "city_locations", default=0)
        if isinstance(cities, (list, tuple)):
            cities = len(cities)
        roads = _safe_get(player_state, "roads", "road_count", default=0)
        if isinstance(roads, (list, tuple)):
            roads = len(roads)
        dev_vp = _safe_get(player_state, "dev_vp", "dev_victory_points", default=0)

        # Resources summary
        resources_obj = _safe_get(player_state, "resources", default=0)
        resources_total = 0
        resource_diversity = 0
        try:
            if isinstance(resources_obj, dict):
                resources_total = sum(resources_obj.values())
                resource_diversity = sum(1 for v in resources_obj.values() if v > 0)
            elif isinstance(resources_obj, (list, tuple)):
                resources_total = sum(resources_obj)
                resource_diversity = sum(1 for v in resources_obj if v > 0)
            else:
                resources_total = int(resources_obj)
                resource_diversity = 1 if resources_total > 0 else 0
        except Exception:
            resources_total = 0
            resource_diversity = 0

        # Production potential estimation
        prod_value = 0.0
        try:
            board = getattr(state, "board", None) or getattr(game, "board", None)
            hexes = getattr(board, "hexes", None) or getattr(board, "tiles", None) or []
            settlements_list = _safe_get(player_state, "settlements", "settle_locations", default=[])
            if isinstance(settlements_list, (list, tuple)):
                for s in settlements_list:
                    try:
                        for h in hexes:
                            neighbors = getattr(h, "vertices", 
                            None) or getattr(h, "adjacent_vertices", 
                            None) or []
                            if s in neighbors:
                                num = getattr(h, "roll", 
                                None) or getattr(h, "number", 
                                None) or getattr(h, "value", 
                                None)
                                try:
                                    num = int(num)
                                except Exception:
                                    num = None
                                if num in die_prob:
                                    prod_value += die_prob[num] * 1.0
                    except Exception:
                        continue
            cities_list = _safe_get(player_state, "cities", "city_locations", default=[])
            if isinstance(cities_list, (list, tuple)):
                for c in cities_list:
                    try:
                        for h in hexes:
                            neighbors = getattr(h, "vertices", 
                            None) or 
                            getattr(h, "adjacent_vertices", None) or []
                            if c in neighbors:
                                num = getattr(h, "roll", 
                                None) or getattr(h, "number", 
                                None) or getattr(h, "value", 
                                None)
                                try:
                                    num = int(num)
                                except Exception:
                                    num = None
                                if num in die_prob:
                                    prod_value += die_prob[num] * 2.0
                    except Exception:
                        continue
        except Exception:
            prod_value = 0.0

        # City upgrade progress heuristic
        city_resource_val = 0.0
        try:
            if isinstance(resources_obj, dict):
                wheat = resources_obj.get("wheat", 0) + resources_obj.get("grain", 0)
                ore = resources_obj.get("ore", 0) + resources_obj.get("metal", 0)
                city_resource_val = min(wheat, ore)
        except Exception:
            city_resource_val = 0.0

        # Phase multipliers
        phase = self.get_game_phase(game, color)
        mults = self.MULTS.get(phase, self.MULTS["MID"])
        settlement_mul = mults["settlement"]
        road_mul = mults["road"]
        city_mul = mults["city"]
        dev_mul = mults["dev"]

        # Adjust production weight by phase
        prod_weight = 80.0 if phase == "EARLY" else 45.0 if phase == "MID" else 30.0

        # Compose weighted sum (city reward scaled by city_mul)
        score = (
            float(vp) * 100.0
            + float(settlements) * 25.0 * settlement_mul
            + float(cities) * 60.0 * city_mul
            + float(roads) * 6.0 * road_mul
            + float(dev_vp) * 50.0
            + float(resources_total) * 1.0
            + float(resource_diversity) * 3.0
            + float(city_resource_val) * 5.0
            + float(prod_value) * prod_weight
        )

        return float(score)

    def _evaluate_game_state(self, game: Game, color: Color) -> float:
        """Evaluate a single game state for the given player color.

        Prefer adapters.base_fn() if available (cached in self._value_fn). If available, combine
        it with the heuristic for stability. We keep phase multipliers inside the heuristic so
        they influence the final blended value.
        """
        heuristic = self._heuristic_value(game, color)
        if self._value_fn is not None:
            try:
                vf_val = float(self._value_fn(game, color))
                return 0.85 * vf_val + 0.15 * heuristic
            except Exception as e:
                self.debug_print("FooPlayer: value_fn failed during evaluate_game_state, falling back to heuristic. Error:", e)
        return float(heuristic)

    # ------------------- Cheap scoring & potentials -------------------
    def _get_player_state(self, game: Game, color: Color) -> Any:
        """Return the player_state object from the game state (best-effort)."""
        try:
            state = getattr(game, "state", game)
            players = getattr(state, "players", None) or getattr(game, "players", None)
            if isinstance(players, dict):
                return players.get(color) or players.get(str(color))
            elif isinstance(players, (list, tuple)):
                for p in players:
                    if getattr(p, "color", None) == color or getattr(p, "color", None) == str(color):
                        return p
        except Exception:
            return None
        return None

    def settlement_potential(self, action: Any, game: Game, color: Color) -> float:
        """Estimate benefit of a settlement action: new resource types and production.

        Best-effort: try to parse adjacent hexes from action or fallback to string heuristics.
        """
        bonus = 0.0
        try:
            name = self._safe_action_name(action)
            # Quick check: if action indicates a settlement, give base
            if any(tok in name for tok in ("build_settlement", "build_sett", "settle")):
                bonus += 5.0

            # Try to parse a vertex index from the action string
            digits = [int(tok) for tok in name.split() if tok.isdigit()]
            vertex = digits[0] if digits else None

            state = getattr(game, "state", game)
            board = getattr(state, "board", None) or getattr(game, "board", None)
            hexes = getattr(board, "hexes", None) or getattr(board, "tiles", None) or []

            # Player's current resource types
            player_state = self._get_player_state(game, color)
            player_types = set()
            try:
                settlements_list = getattr(player_state, "settlements", None) or getattr(player_state, "settle_locations", None) or []
                if isinstance(settlements_list, (list, tuple)):
                    for s in settlements_list:
                        for h in hexes:
                            neighbors = getattr(h, "vertices", 
                            None) or getattr(h, "adjacent_vertices", 
                            None) or []
                            if s in neighbors:
                                rtype = getattr(h, "resource", 
                                None) or getattr(h, "type", 
                                None)
                                if rtype is not None:
                                    player_types.add(str(rtype).lower())
            except Exception:
                player_types = set()

            # Adjacent resources for proposed vertex
            adj_resources = set()
            prod_sum = 0.0
            die_prob = {2: 1 / 36, 3: 2 / 36, 4: 3 / 36, 5: 4 / 36, 6: 5 / 36, 8: 5 / 36, 9: 4 / 36, 10: 3 / 36, 11: 2 / 36, 12: 1 / 36}
            if vertex is not None:
                for h in hexes:
                    try:
                        neighbors = getattr(h, "vertices", 
                        None) or getattr(h, "adjacent_vertices", 
                        None) or []
                        if vertex in neighbors:
                            rtype = getattr(h, "resource", 
                            None) or getattr(h, "type", 
                            None)
                            if rtype is not None:
                                adj_resources.add(str(rtype).lower())
                            num = getattr(h, "roll", 
                            None) or getattr(h, "number", 
                            None) or getattr(h, "value", 
                            None)
                            try:
                                num = int(num)
                            except Exception:
                                num = None
                            if num in die_prob:
                                prod_sum += die_prob[num]
                    except Exception:
                        continue
            # New types
            new_types = adj_resources - player_types
            bonus += float(len(new_types)) * 12.0
            bonus += float(prod_sum) * 8.0
        except Exception:
            pass
        return float(bonus)

    def road_connection_potential(self, action: Any, game: Game, color: Color) -> float:
        """Estimate if a road action helps expansion. Best-effort using indices."""
        bonus = 0.0
        try:
            name = self._safe_action_name(action)
            # try to extract numbers from action name
            digits = [int(tok) for tok in name.split() if tok.isdigit()]
            # player's settlement/city vertices
            player_state = self._get_player_state(game, color)
            player_nodes = set()
            try:
                settles = getattr(player_state, "settlements", None) or getattr(player_state, "settle_locations", None) or []
                cities = getattr(player_state, "cities", None) or getattr(player_state, "city_locations", None) or []
                if isinstance(settles, (list, tuple)):
                    player_nodes.update(settles)
                if isinstance(cities, (list, tuple)):
                    player_nodes.update(cities)
            except Exception:
                player_nodes = set()

            if digits:
                # if any digit matches a player node, give higher bonus
                if any(d in player_nodes for d in digits):
                    bonus += 6.0
                else:
                    bonus += 3.0
            else:
                # fallback string heuristics
                if "build_road" in name or ("road" in name and "build" in name):
                    bonus += 2.0
        except Exception:
            pass
        return float(bonus)

    def evaluate_buy_dev_card(self, action: Any, game: Game, color: Color) -> bool:
        """Decide whether buying a dev card is currently a good idea (best-effort)."""
        try:
            player_state = self._get_player_state(game, color)
            resources = getattr(player_state, "resources", None)
            if isinstance(resources, dict):
                ore = resources.get("ore", 0) + resources.get("metal", 0)
                wheat = resources.get("wheat", 0) + resources.get("grain", 0)
                others = sum(v for k, v in resources.items() if k not in ("ore", "metal", "wheat", "grain"))
                # if have ore+wheat+another, prefer dev card; or if no settlement/road/city affordable
                if ore >= 1 and wheat >= 1 and others >= 1:
                    return True
                # fallback: if early game and we have some resources but no settlement potential, allow dev buy
                phase = self.get_game_phase(game, color)
                if phase == "EARLY" and (ore + wheat + others) >= 3:
                    return True
        except Exception:
            pass
        return False

    def dev_card_ev_estimate(self, game: Game, color: Color) -> float:
        """Estimate expected VP-equivalent value of buying a development card.

        Uses static DEV_DECK and EV_* constants and scales by opponent pressure and army gaps.
        Returns a small VP-equivalent number (e.g., ~0.3-0.6 when favorable).
        """
        try:
            base_ev = 0.0
            # composition-based base EV
            base_ev += (self.DEV_DECK.get("knight", 0) / self.DEV_TOTAL) * self.EV_KNIGHT
            base_ev += (self.DEV_DECK.get("vp", 0) / self.DEV_TOTAL) * self.EV_VP
            base_ev += (self.DEV_DECK.get("road_building", 0) / self.DEV_TOTAL) * self.EV_ROAD_BUILDING
            base_ev += (self.DEV_DECK.get("year_of_plenty", 0) / self.DEV_TOTAL) * self.EV_YOP
            base_ev += (self.DEV_DECK.get("monopoly", 0) / self.DEV_TOTAL) * self.EV_MONOPOLY

            # Scale factors: opponents production pressure and army proximity
            # Compute opponents' max production (best-effort)
            state = getattr(game, "state", game)
            board = getattr(state, "board", None) or getattr(game, "board", None)
            hexes = getattr(board, "hexes", None) or getattr(board, "tiles", None) or []

            opponents = []
            players = getattr(state, "players", None) or getattr(game, "players", None) or []
            my_color = color
            if isinstance(players, dict):
                for k, p in players.items():
                    if k == my_color or getattr(p, "color", None) == my_color:
                        continue
                    opponents.append(p)
            else:
                for p in players:
                    if getattr(p, "color", None) == my_color:
                        continue
                    opponents.append(p)

            # compute simple production score for each opponent
            die_prob = {2: 1 / 36, 3: 2 / 36, 4: 3 / 36, 5: 4 / 36, 6: 5 / 36, 8: 5 / 36, 9: 4 / 36, 10: 3 / 36, 11: 2 / 36, 12: 1 / 36}
            max_opp_prod = 0.0
            for opp in opponents:
                prod = 0.0
                opp_settles = getattr(opp, "settlements", None) or getattr(opp, "settle_locations", None) or []
                opp_cities = getattr(opp, "cities", None) or getattr(opp, "city_locations", None) or []
                try:
                    for s in opp_settles:
                        for h in hexes:
                            neighbors = getattr(h, "vertices", 
                            None) or getattr(h, "adjacent_vertices", 
                            None) or []
                            if s in neighbors:
                                num = getattr(h, "roll", 
                                None) or getattr(h, "number", 
                                None) or getattr(h, "value", 
                                None)
                                try:
                                    num = int(num)
                                except Exception:
                                    num = None
                                if num in die_prob:
                                    prod += die_prob[num]
                    for c in opp_cities:
                        for h in hexes:
                            neighbors = getattr(h, "vertices", 
                            None) or getattr(h, "adjacent_vertices", 
                            None) or []
                            if c in neighbors:
                                num = getattr(h, "roll", 
                                None) or getattr(h, "number", 
                                None) or getattr(h, "value", 
                                None)
                                try:
                                    num = int(num)
                                except Exception:
                                    num = None
                                if num in die_prob:
                                    prod += 2.0 * die_prob[num]
                except Exception:
                    pass
                max_opp_prod = max(max_opp_prod, prod)

            # army gap factor
            my_state = self._get_player_state(game, color)
            my_army = getattr(my_state, "army", None) or getattr(my_state, "army_size", None) or getattr(my_state, "knights_played", None) or 0
            try:
                my_army = int(my_army)
            except Exception:
                my_army = 0
            max_other_army = 0
            try:
                if isinstance(players, dict):
                    for k, p in players.items():
                        if k == my_color or getattr(p, "color", None) == my_color:
                            continue
                        oa = getattr(p, "army", None) or 
                        getattr(p, "army_size", None) or 
                        getattr(p, "knights_played", 
                        None) or 0
                        try:
                            oa = int(oa)
                        except Exception:
                            oa = 0
                        max_other_army = max(max_other_army, oa)
                else:
                    for p in players:
                        if getattr(p, "color", None) == my_color:
                            continue
                        oa = getattr(p, "army", None) or 
                        getattr(p, "army_size", None) or 
                        getattr(p, "knights_played", None) or 0
                        try:
                            oa = int(oa)
                        except Exception:
                            oa = 0
                        max_other_army = max(max_other_army, oa)
            except Exception:
                max_other_army = 0

            army_gap = max(0, max_other_army - my_army)

            # scale base_ev conservatively
            scale = 1.0
            if max_opp_prod > 0.25:  # opponent has strong production
                scale += 0.25
            if army_gap >= 1:
                scale += 0.15 * army_gap

            final_ev = base_ev * scale
            return float(final_ev)
        except Exception:
            # fallback conservative
            return 0.25

    def build_urgency(self, game: Game, color: Color) -> Tuple[float, float, float]:
        """Return (city_bonus, settlement_bonus, road_bonus) depending on resources and phase."""
        city_bonus = 0.0
        settlement_bonus = 0.0
        road_bonus = 0.0
        try:
            player_state = self._get_player_state(game, color)
            resources = getattr(player_state, "resources", None) or {}
            if not isinstance(resources, dict):
                # try to coerce
                try:
                    total = sum(resources)
                    resources = {"res": total}
                except Exception:
                    resources = {}

            # simple can_afford_city_soon heuristic
            ore = resources.get("ore", 0) + resources.get("metal", 0)
            wheat = resources.get("wheat", 0) + resources.get("grain", 0)
            settlements_list = getattr(player_state, "settlements", None) or getattr(player_state, "settle_locations", None) or []
            settlements_owned = len(settlements_list) if isinstance(settlements_list, (list, tuple)) else 0

            phase = self.get_game_phase(game, color)
            # If mid/late and can afford city soon, large city urgency
            if phase in ("MID", "LATE") and ore >= 2 and wheat >= 1:
                city_bonus += 40.0
                self._diag_city_urgency_count += 1
            # If early and lacking settlements target, encourage settlements strongly
            if phase == "EARLY" and settlements_owned < self.TARGET_SETTLEMENTS_EARLY:
                settlement_bonus += 35.0
                self._diag_settle_urgency_count += 1
            # Road potential: give moderate constant bonus
            road_bonus += 10.0
        except Exception:
            pass
        return city_bonus, settlement_bonus, road_bonus

    def cheap_pre_score(self, action: Any, game: Game, color: Color) -> float:
        """Cheap, fast scoring used to prioritize actions for simulation (phase-aware)."""
        s = 0.0
        name = self._safe_action_name(action)

        phase = self.get_game_phase(game, color)
        mults = self.MULTS.get(phase, self.MULTS["MID"])
        settlement_mul = mults["settlement"]
        road_mul = mults["road"]
        city_mul = mults["city"]
        dev_mul = mults["dev"]

        # urgency bonuses
        city_urgency, sett_urgency, road_urgency = self.build_urgency(game, color)

        # Reward direct VP gains but adjust city bias early
        if any(tok in name for tok in ("build_city",)):
            base_city = max(50.0, 100.0 * city_mul - 15.0)
            # penalize city if early and still below settlement target
            try:
                player_state = self._get_player_state(game, color)
                settles = getattr(player_state, "settlements", None) or getattr(player_state, "settle_locations", None) or []
                curr_settlements = len(settles) if isinstance(settles, (list, tuple)) else 0
                if phase == "EARLY" and curr_settlements < 
                self.TARGET_SETTLEMENTS_EARLY:
                    base_city *= 0.6
            except Exception:
                pass
            s += base_city + city_urgency

        if any(tok in name for tok in ("build_settlement", "build_sett")):
            s += 90.0 * settlement_mul
            # add settlement potential (resource diversity / production)
            s += self.settlement_potential(action, game, color) * (1.0 if phase != "EARLY" else settlement_mul)
            s += sett_urgency

        if "buy_dev" in name or "buycard" in name or "buy_dev_card" in name:
            # compute EV estimate
            dev_ev = self.dev_card_ev_estimate(game, color)
            s += dev_ev * self.DEV_EV_SCALE
            # slightly reduced base bias to favor cities when urgent
            if self.evaluate_buy_dev_card(action, game, color):
                s += 8.0 * dev_mul
            try:
                if dev_ev >= self.DEV_EV_THRESHOLD:
                    s += 2.0
            except Exception:
                pass

        if "build_road" in name or ("road" in name and "build" in name):
            s += 20.0 * road_mul
            s += self.road_connection_potential(action, game, color) * (1.0 if phase != "EARLY" else road_mul)
            s += road_urgency

        if "knight" in name or "play_knight" in name:
            # raise baseline and include army/steal bonuses
            s += 70.0
            s += self.evaluate_play_knight(action, game, color)

        if "robber" in name or "move_robber" in name:
            s += 50.0
            s += self.evaluate_robber_action(action, game, color)

        if "trade" in name or "offer_trade" in name:
            s += 10.0

        # Encourage hitting settlement target early
        try:
            player_state = self._get_player_state(game, color)
            curr_settlements = 0
            settles = getattr(player_state, "settlements", None) or getattr(player_state, "settle_locations", None) or []
            if isinstance(settles, (list, tuple)):
                curr_settlements = len(settles)
            if phase == "EARLY" and curr_settlements < self.TARGET_SETTLEMENTS_EARLY and any(tok in name for tok in ("build_settlement", "build_sett")):
                s += 30.0
        except Exception:
            pass

        # small settlement/road potentials for other actions
        if not any(tok in name for tok in ("build_settlement", "build_sett")):
            s += self.settlement_potential(action, game, color) * 0.1
        if not any(tok in name for tok in ("build_road",)):
            s += self.road_connection_potential(action, game, color) * 0.1

        # Minor random tie-break
        s += random.random() * 1e-3
        return s

    # ------------------- Prefilter actions (phase-aware guarantees) -------------------
    def prefilter_actions(self, actions: List[Any], game: Game, color: Color) -> List[Any]:
        """Return a bounded list of candidate actions to evaluate thoroughly.

        Guarantees inclusion of must-include tokens and early-game settlement/road actions.
        """
        if not actions:
            return []

        all_actions = list(actions)
        phase = self.get_game_phase(game, color)

        musts = []
        others = []
        found_settlement = None
        found_road = None
        for a in all_actions:
            name = self._safe_action_name(a)
            if any(tok in name for tok in self.MUST_INCLUDE_TOKENS):
                if a not in musts:
                    musts.append(a)
            else:
                others.append(a)
            if found_settlement is None and any(tok in name for tok in ("build_settlement", "build_sett", "settle")):
                found_settlement = a
            if found_road is None and any(tok in name for tok in ("build_road", "road")):
                found_road = a

        # Phase-based forced includes: ensure at least one settlement and one road action if present in EARLY
        if phase == "EARLY":
            if found_settlement is not None and found_settlement not in musts:
                musts.append(found_settlement)
                self._diag_forced_settlement += 1
            if found_road is not None and found_road not in musts:
                musts.append(found_road)
                self._diag_forced_road += 1

        # Include recommended dev-card buys if conservative and EV threshold met
        for a in all_actions:
            name = self._safe_action_name(a)
            if any(tok in name for tok in ("buy_dev", "buycard", "buy_dev_card")):
                try:
                    if self.evaluate_buy_dev_card(a, game, color):
                        dev_ev = self.dev_card_ev_estimate(game, color)
                        if dev_ev >= self.DEV_EV_THRESHOLD and a not in musts:
                            # include only if dev EV merits it
                            musts.append(a)
                except Exception:
                    pass

        # Ensure robber/knight actions are present
        for a in all_actions:
            name = self._safe_action_name(a)
            if any(tok in name for tok in ("robber", "move_robber", "knight", "play_knight")):
                if a not in musts:
                    musts.append(a)

        # Score and pick top-K from others
        scored = [(self.cheap_pre_score(a, game, color), a) for a in others]
        scored.sort(key=lambda x: x[0], reverse=True)
        top_k = [a for (_s, a) in scored[: self.PREFILTER_TOP_K]]

        # Combine unique musts + top_k preserving order
        candidates = []
        for a in musts + top_k:
            if a not in candidates:
                candidates.append(a)

        # Fill up with random remaining samples until MAX_SIMULATIONS
        remaining = [a for a in all_actions if a not in candidates]
        random.shuffle(remaining)
        while len(candidates) < min(len(all_actions), self.MAX_SIMULATIONS) and remaining:
            candidates.append(remaining.pop())

        if not candidates and all_actions:
            candidates = random.sample(all_actions, min(len(all_actions), self.MAX_SIMULATIONS))

        self.debug_print(f"FooPlayer: Prefilter selected {len(candidates)} candidates (musts={len(musts)}, phase={phase})")
        if self.DEBUG and phase == "EARLY":
            self.debug_print(f"  Forced includes: settlement={'yes' if found_settlement else 'no'}, road={'yes' if found_road else 'no'}")
        return candidates

    # ------------------- Playable actions extraction -------------------
    def get_playable_actions_from_game(self, game: Game) -> List[Any]:
        """Try adapters.list_prunned_actions first, then common game attributes."""
        try:
            acts = list_prunned_actions(game)
            if acts:
                return acts
        except Exception as e:
            self.debug_print("FooPlayer: list_prunned_actions unavailable or failed. Error:", e)

        try:
            if hasattr(game, "get_playable_actions"):
                return list(game.get_playable_actions())
        except Exception:
            pass
        try:
            if hasattr(game, "playable_actions"):
                return list(getattr(game, "playable_actions"))
        except Exception:
            pass
        try:
            state = getattr(game, "state", None)
            if state is not None and hasattr(state, "playable_actions"):
                return list(getattr(state, "playable_actions"))
        except Exception:
            pass

        return []

    # ------------------- Robber / Knight evaluation -------------------
    def evaluate_robber_action(self, action: Any, game: Game, color: Color) -> float:
        """Estimate the value of moving the robber (best-effort).

        If the action does not specify a target hex, evaluate all hexes and prefer the
        one that maximizes opponent production loss.
        """
        score = 0.0
        try:
            # Base preference to include robber moves (use HIGH base for aggressive play)
            score += self.ROBBER_BASE_SCORE_HIGH
            name = self._safe_action_name(action)
            # Try to parse a target hex id
            digits = [int(tok) for tok in name.split() if tok.isdigit()]
            target = digits[0] if digits else None

            # Die probabilities
            die_prob = {2: 1 / 36, 3: 2 / 36, 4: 3 / 36, 5: 4 / 36, 6: 5 / 36, 8: 5 / 36, 9: 4 / 36, 10: 3 / 36, 11: 2 / 36, 12: 1 / 36}

            state = getattr(game, "state", game)
            board = getattr(state, "board", None) or getattr(game, "board", None)
            hexes = getattr(board, "hexes", None) or getattr(board, "tiles", None) or []

            # Map hex identifier to object (best-effort: use index or id)
            hex_map = {}
            for idx, h in enumerate(hexes):
                try:
                    hid = getattr(h, "id", None) or getattr(h, "index", None) or idx
                except Exception:
                    hid = idx
                try:
                    key = int(hid) if isinstance(hid, int) or (isinstance(hid, str) and hid.isdigit()) else idx
                except Exception:
                    key = idx
                hex_map[key] = h

            # Determine best target if none specified
            targets_to_consider = [target] if target in hex_map else list(hex_map.keys())

            # Compute production loss on opponents per candidate target
            opponents = []
            players = getattr(state, "players", None) or getattr(game, "players", None) or []
            my_color = color
            if isinstance(players, dict):
                for k, p in players.items():
                    if k == my_color or getattr(p, "color", None) == my_color:
                        continue
                    opponents.append(p)
            else:
                for p in players:
                    if getattr(p, "color", None) == my_color:
                        continue
                    opponents.append(p)

            best_loss = 0.0
            best_steal = 0.0
            best_hex = None
            resource_value = {"ore": 3.0, "metal": 3.0, "wheat": 3.0, "grain": 3.0, "brick": 2.0, "lumber": 2.0, "wood": 2.0, "sheep": 2.0}

            for t in targets_to_consider:
                try:
                    if t not in hex_map:
                        continue
                    h = hex_map[t]
                    num = getattr(h, "roll", None) or getattr(h, "number", None) or getattr(h, "value", None)
                    try:
                        num = int(num)
                    except Exception:
                        num = None
                    prob = die_prob.get(num, 0)
                    total_prod_loss = 0.0
                    steal_expected = 0.0
                    for opp in opponents:
                        
                        opp_settles = getattr(opp, 
                        "settlements", None) or getattr(opp, 
                        "settle_locations", None) or []
                        opp_cities = getattr(opp, "cities", 
                        None) or getattr(opp, "city_locations", 
                        None) or []
                        mult = 0.0
                        try:
                            for s in opp_settles:
                                neighbors = getattr(h, 
                                "vertices", None) or getattr(h, 
                                "adjacent_vertices", None) or []
                                if s in neighbors:
                                    mult += 1.0
                            for c in opp_cities:
                                neighbors = getattr(h, 
                                "vertices", None) or getattr(h, 
                                "adjacent_vertices", None) or []
                                if c in neighbors:
                                    mult += 2.0
                        except Exception:
                            continue
                        total_prod_loss += prob * mult
                        # Estimate steal expected
                        try:
                            opp_resources = getattr(opp, 
                            "resources", None) or {}
                            if isinstance(opp_resources, dict) 
                            and opp_resources:
                                total_res = 
                                sum(opp_resources.values())
                                if total_res > 0:
                                    avg_val = 
                                    sum(resource_value.get(r, 
                                    1.5) * (opp_resources.get(r, 
                                    0) / total_res) for r in 
                                    opp_resources)
                                    steal_expected += avg_val * 
                                    0.5
                        except Exception:
                            pass
                    # choose best
                    if total_prod_loss > best_loss or (abs(total_prod_loss - best_loss) < 1e-9 and steal_expected > best_steal):
                        best_loss = total_prod_loss
                        best_steal = steal_expected
                        best_hex = t
                except Exception:
                    continue

            # Aggressive scaling per latest tuning
            score += best_loss * self.PROD_LOSS_IMPORTANCE
            score += best_steal * 30.0
            # Extra bonus if multiple opponent cities affected
            try:
                if best_hex in hex_map:
                    h = hex_map[best_hex]
                    city_count = 0
                    for opp in opponents:
                        for c in getattr(opp, "cities", []) or 
                        getattr(opp, "city_locations", []) or []:
                            neighbors = getattr(h, "vertices", 
                            None) or getattr(h, 
                            "adjacent_vertices", None) or []
                            if c in neighbors:
                                city_count += 1
                    if city_count > 0:
                        score += 20.0 * city_count
            except Exception:
                pass

            # If steal estimated is very significant, add 
            decisive bonus
            if best_steal > 2.0:
                score += 30.0

            # Debug
            if self.DEBUG and best_hex is not None:
                self.debug_print(f"FooPlayer: evaluate_robber_action best_hex={best_hex} prod_loss={best_loss:.3f} steal_ev={best_steal:.2f}")

        except Exception:
            pass
        return float(score)

    def evaluate_play_knight(self, action: Any, game: Game, color: Color) -> float:
        """Estimate the value of playing a knight (best-effort)."""
        score = float(self.KNIGHT_BASE)
        try:
            name = self._safe_action_name(action)
            if "steal" in name or "rob" in name:
                score += 10.0

            # army progress
            player_state = self._get_player_state(game, color)
            army = getattr(player_state, "army", None) or getattr(player_state, "army_size", None) or getattr(player_state, "knights_played", None) or 0
            try:
                army = int(army)
            except Exception:
                army = 0

            # detect largest army threshold
            largest_threshold = 3
            try:
                state = getattr(game, "state", game)
                players = getattr(state, "players", None) or getattr(game, "players", None) or []
                max_other = 0
                if isinstance(players, dict):
                    for k, p in players.items():
                        if getattr(p, "color", None) == color or k == color:
                            continue
                        other_army = getattr(p, "army", None) or getattr(p, "army_size", None) or getattr(p, "knights_played", None) or 0
                        try:
                            other_army = int(other_army)
                        except Exception:
                            other_army = 0
                        max_other = max(max_other, other_army)
                else:
                    for p in players:
                        if getattr(p, "color", None) == color:
                            continue
                        other_army = getattr(p, "army", None) or getattr(p, "army_size", None) or getattr(p, "knights_played", None) or 0
                        try:
                            other_army = int(other_army)
                        except Exception:
                            other_army = 0
                        max_other = max(max_other, other_army)
                largest_threshold = max(3, max_other + 1)
            except Exception:
                largest_threshold = 3

            if army + 1 >= largest_threshold:
                score += self.KNIGHT_LARGEST_ARMY_BONUS
            else:
                score += 20.0

            # Debug
            if self.DEBUG:
                self.debug_print(f"FooPlayer: evaluate_play_knight army={army} target={largest_threshold} score={score}")
        except Exception:
            pass
        return float(score)

    # ------------------- Helper: determine active player color -------------------
    def _get_active_player_color(self, game: Game) -> Optional[Color]:
        """Best-effort to detect which Color is to move in the given game state."""
        try:
            state = getattr(game, "state", game)
            cp = getattr(state, "current_player", None) or getattr(state, "active_player", None) or getattr(state, "turn_color", None)
            if cp is None:
                cp = getattr(game, "current_player", None)
            # cp might be index, player object, or Color
            if isinstance(cp, Color):
                return cp
            if isinstance(cp, int):
                players = getattr(state, "players", None) or getattr(game, "players", None) or []
                try:
                    if isinstance(players, (list, tuple)) and 0 <= cp < len(players):
                        return getattr(players[cp], "color", None)
                except Exception:
                    pass
            # If cp is a player object
            if hasattr(cp, "color"):
                return getattr(cp, "color")

            # Fallback: pick first player in players whose color != our color
            players = getattr(state, "players", None) or getattr(game, "players", None) or []
            my_color = self._get_player_color()
            if isinstance(players, dict):
                for k, p in players.items():
                    try:
                        c = getattr(p, "color", None) or k
                        if c != my_color:
                            return c
                    except Exception:
                        continue
            else:
                for p in players:
                    try:
                        c = getattr(p, "color", None)
                        if c != my_color:
                            return c
                    except Exception:
                        continue
        except Exception:
            pass
        return None

    # ------------------- Rollout logic with opponent-response -------------------
    def rollout_value(self, game: Game, color: Color, depth: int, initial: bool = True) -> float:
        """Short greedy rollout with phase bias and light opponent-response.

        initial: True for the first step of rollout so we can bias toward expansion early.
        """
        try:
            if depth <= 0:
                return self._evaluate_game_state(game, color)

            actions = self.get_playable_actions_from_game(game)
            if not actions:
                return self._evaluate_game_state(game, color)

            phase = self.get_game_phase(game, color)

            def score_for_rollout(a, g, c, is_initial):
                base = self.cheap_pre_score(a, g, c)
                if is_initial and phase == "EARLY":
                    name = self._safe_action_name(a)
                    if any(tok in name for tok in ("build_settlement", "build_sett", "settle")):
                        base *= self.ROLLOUT_SETTLEMENT_BONUS
                    if any(tok in name for tok in ("build_road", "road")):
                        base *= self.ROLLOUT_ROAD_BONUS
                return base

            sorted_actions = sorted(actions, key=lambda a: score_for_rollout(a, game, color, initial), reverse=True)

            # Try top actions to simulate
            for a in sorted_actions[:6]:
                branches = []
                try:
                    branches = execute_deterministic(game, a)
                except Exception:
                    try:
                        branches = execute_spectrum(game, a)
                    except Exception:
                        branches = []
                if not branches:
                    continue
                # pick the most probable branch
                next_game = max(branches, key=lambda bp: float(bp[1]))[0]

                # Light opponent-response: if opponent to move next, simulate their greedy action once
                opp_color = self._get_active_player_color(next_game)
                my_color = color
                if opp_color is not None and opp_color != my_color and depth >= 2:
                    try:
                        opp_actions = self.get_playable_actions_from_game(next_game)
                        if opp_actions:
                            # filter out robber/knight for 
                            opponent response unless all are 
                            robber/knight
                            non_disrupt = [oa for oa in 
                            opp_actions if not any(tok in 
                            self._safe_action_name(oa) for tok 
                            in ("knight", "robber", 
                            "move_robber"))]
                            candidate_ops = non_disrupt if 
                            non_disrupt else opp_actions
                            # pick opponent best action by 
                            cheap_pre_score from their 
                            perspective
                            best_opp = max(candidate_ops, 
                            key=lambda oa: 
                            self.cheap_pre_score(oa, next_game, 
                            opp_color))
                            # simulate opponent action 
                            deterministically if possible
                            opp_branches = []
                            try:
                                opp_branches = 
                                execute_deterministic(next_game, 
                                best_opp)
                            except Exception:
                                try:
                                    opp_branches = 
                                    execute_spectrum(next_game, 
                                    best_opp)
                                except Exception:
                                    opp_branches = []
                            if opp_branches:
                                next_game = max(opp_branches, 
                                key=lambda bp: float(bp[1]))[0]
                    except Exception:
                        pass

                return self.rollout_value(next_game, color, 
                depth - 1, initial=False)

            # fallback: try any action that simulates
            for a in sorted_actions[:10]:
                branches = []
                try:
                    branches = execute_deterministic(game, a)
                except Exception:
                    try:
                        branches = execute_spectrum(game, a)
                    except Exception:
                        branches = []
                if branches:
                    next_game = max(branches, key=lambda bp: 
                    float(bp[1]))[0]
                    return self.rollout_value(next_game, color, 
                    depth - 1, initial=False)

            return self._evaluate_game_state(game, color)
        except Exception as e:
            self.debug_print("FooPlayer: rollout_value exception, falling back to evaluate_game_state. Error:", e)
            return self._evaluate_game_state(game, color)

    # ------------------- Evaluate action expectation (enhanced) -------------------
    def _evaluate_action_expectation(self, game: Game, action: Any, per_action_branch_limit: int = 8) -> float:
        """Compute expected value of taking `action` in `game` for this player.

        Uses execute_spectrum when available then adds a rollout estimate for depth-1.
        """
        color = self._get_player_color()

        # Quick boosts for robber/knight/dev before heavy sim
        name = self._safe_action_name(action)
        preboost = 0.0
        try:
            if any(tok in name for tok in ("move_robber", "robber")):
                preboost += self.evaluate_robber_action(action, game, color)
            if any(tok in name for tok in ("knight", "play_knight")):
                preboost += self.evaluate_play_knight(action, game, color)
            if any(tok in name for tok in ("buy_dev", "buycard", "buy_dev_card")):
                try:
                    dev_ev = self.dev_card_ev_estimate(game, color)
                    preboost += dev_ev * self.DEV_EV_SCALE
                except Exception:
                    # fallback small preboost
                    preboost += 20.0
        except Exception:
            preboost += 0.0

        branches = None
        try:
            branches = execute_spectrum(game, action)
            if not branches:
                raise RuntimeError("execute_spectrum returned no branches")
        except Exception as e_s:
            self.debug_print("FooPlayer: execute_spectrum failed or unavailable for action; trying deterministic. Error:", e_s)
            try:
                branches = execute_deterministic(game, action)
                if not branches:
                    raise RuntimeError("execute_deterministic returned no outcomes")
            except Exception as e_d:
                self.debug_print("FooPlayer: Both execute_spectrum and execute_deterministic failed for action. Errors:", e_s, e_d)
                return float("-inf")

        # Limit branches to keep runtime bounded
        if len(branches) > per_action_branch_limit:
            branches = sorted(branches, key=lambda bp: float(bp[1]), reverse=True)[:per_action_branch_limit]

        expected = 0.0
        total_prob = 0.0
        rollout_depth = max(0, self.ROLLOUT_DEPTH - 1)
        for (out_game, prob) in branches:
            try:
                # For buy_dev actions, if the branch encodes a known draw outcome, we could refine.
                # In absence of explicit draw info, rely on dev_ev_estimate as a conservative proxy.
                immediate = self._evaluate_game_state(out_game, color)
                rollout_est = self.rollout_value(out_game, color, rollout_depth, initial=True)
                branch_val = 0.6 * immediate + 0.4 * rollout_est
            except Exception as e:
                self.debug_print("FooPlayer: evaluation failed for branch, using heuristic. Error:", e)
                branch_val = self._heuristic_value(out_game, color)
            expected += float(prob) * float(branch_val)
            total_prob += float(prob)

        if total_prob > 0:
            expected = expected / total_prob

        expected += preboost
        return float(expected)

    # ------------------- Main decision function -------------------
    def decide(self, game: Game, playable_actions: Iterable) -> Optional[object]:
        """Choose an action from playable_actions using phase-aware sampling + rollouts."""
        try:
            playable_actions = list(playable_actions)
            if not playable_actions:
                self.debug_print("FooPlayer: No playable actions available, returning None")
                return None

            color = self._get_player_color()
            phase = self.get_game_phase(game, color)

            # Prefilter candidate actions
            candidates = self.prefilter_actions(playable_actions, game, color)

            # Cap to MAX_SIMULATIONS
            if len(candidates) > self.MAX_SIMULATIONS:
                candidates = candidates[: self.MAX_SIMULATIONS]

            if not candidates:
                candidates = random.sample(playable_actions, 
                min(len(playable_actions), self.MAX_SIMULATIONS))

            # Distribute simulation budget adaptively
            per_action_budget = max(1, self.SIMULATION_BUDGET // 
            max(1, len(candidates)))

            best_score = float("-inf")
            best_actions: List[Any] = []
            scores_debug: List[Tuple[float, Any]] = []

            for a in candidates:
                try:
                    score = 
                    self._evaluate_action_expectation(game, a, 
                    per_action_branch_limit=per_action_budget)
                except Exception as e:
                    self.debug_print("FooPlayer: Exception 
                    during action evaluation, skipping action. 
                    Error:", e)
                    score = float("-inf")

                scores_debug.append((score, a))

                if score > best_score + self.TOLERANCE:
                    best_score = score
                    best_actions = [a]
                elif abs(score - best_score) <= self.TOLERANCE:
                    best_actions.append(a)

            # If no action had a finite score, fallback to first playable action
            if not best_actions:
                self.debug_print("FooPlayer: All evaluations failed, defaulting to first playable action")
                return playable_actions[0]

            # Epsilon-greedy randomness to reduce predictability
            chosen: Any
            scores_debug.sort(key=lambda x: x[0], reverse=True)
            if random.random() < self.EPSILON_GREEDY and len(scores_debug) >= 2:
                # pick from top-3 weighted by score (or fewer if not available)
                top_k = scores_debug[: min(3, len(scores_debug))]
                weights = [max(0.0, s - top_k[-1][0] + 1e-6) for (s, a) in top_k]
                total_w = sum(weights)
                if total_w > 0:
                    r = random.random() * total_w
                    cum = 0.0
                    for w, (_s, a) in zip(weights, top_k):
                        cum += w
                        if r <= cum:
                            chosen = a
                            break
                    else:
                        chosen = top_k[0][1]
                else:
                    chosen = scores_debug[0][1]
                if self.DEBUG:
                    self.debug_print(f"FooPlayer: EPSILON pick 
                    triggered, chosen alternate action {chosen}")
                return chosen

            # If tie, break ties preferring 
            settlement/road/resource diversity improvements
            if len(best_actions) > 1:
                tie_metrics = []
                for a in best_actions:
                    try:
                        metric = 0.0
                        metric += self.settlement_potential(a,
                        
                        game, color)
                        metric += 
                        self.road_connection_potential(a, game, 
                        color)
                        # small production proxy via heuristic
                        metric += 0.01 * 
                        self._heuristic_value(game, color)
                        tie_metrics.append((metric, a))
                    except Exception:
                        tie_metrics.append((0.0, a))
                tie_metrics.sort(key=lambda x: x[0], 
                reverse=True)
                # pick the top metric actions (could still be 
                multiple)
                top_metric = tie_metrics[0][0]
                filtered = [a for (m, a) in tie_metrics if abs(m 
                - top_metric) <= self.TOLERANCE]
                if filtered:
                    chosen = random.choice(filtered)
                else:
                    chosen = random.choice(best_actions)
            else:
                chosen = best_actions[0]

            # Debug logging: phase and top candidates
            if self.DEBUG:
                self.debug_print(f"FooPlayer: Phase={phase}, 
                SettlementsTarget=
                {self.TARGET_SETTLEMENTS_EARLY}")
                topn = scores_debug[:3]
                self.debug_print("FooPlayer: Top candidates:")
                for sc, act in topn:
                    self.debug_print(f"  score={sc:.2f} action=
                    {act}")

            self.debug_print(f"FooPlayer: Chosen action {chosen} 
            with expected score {best_score}")
            return chosen
        except Exception as e:
            # Protect against unexpected errors
            print("FooPlayer: Unexpected error in decide(), 
            defaulting to first playable action. Error:", e)
            try:
                return list(playable_actions)[0]
            except Exception:
                return None

\end{python}

\subsection{HexMachina (No Discovery) Shallow Strategy}
\begin{python}

import random
import statistics
from .adapters import (
    Game, Player, Color
)

# Hyperparameters for this player. Tweak across evolutions.
K_ROLLOUTS = 0  # rollouts disabled in this adapter-limited implementation
MAX_ROLLOUT_DEPTH = 10  # not used currently; kept for future use
MAX_ACTIONS_TO_EVALUATE = 12
DEBUG = True

class FooPlayer(Player):
    """A stronger FooPlayer that performs a 1-ply lookahead and evaluates
    the immediate successor state using a robust, defensive static evaluator.

    Notes on integration with adapters.py:
    - We only use the thin adapter surface exported above (Game, Player, Color).
    - We call game.copy() to create hypothetical states and game.execute(action)
      to apply actions to those copies. We avoid calling any non-exported
      adapter helpers so this file remains compatible with the framework.

    Limitations and rationale:
    - The adapters surface available in this environment does not explicitly
      expose helper functions for enumerating playable actions from an
      arbitrary game object (those are provided to decide() by the harness).
      Because of this we cannot reliably perform multi-step random rollouts
      (we cannot ask the engine for "playable_actions" inside the player for
      subsequent turns). Attempting to call hypothetical internal APIs would
      risk using non-portable / unsupported functions.
    - To still fix the key flaw (always pick the first action) we implement a
      1-ply lookahead over a sampled set of candidate actions and evaluate the
      successor state with a robust static value function that inspects the
      game.state. This is a significant upgrade over the previous behavior
      and provides a solid foundation for future rollout-based evolution.
    """

    def __init__(self, name=None):
        super().__init__(Color.BLUE, name)

    def decide(self, game, playable_actions):
        """Choose an action from playable_actions.

        Strategy implemented:
        - If there are many playable actions, randomly sample up to
          MAX_ACTIONS_TO_EVALUATE actions to limit computation.
        - For each candidate action, copy the game, execute the action on the
          copy, and evaluate the resulting state with _evaluate_state().
        - Choose the action with the highest evaluation. Break ties randomly.

        The evaluation is defensive: it attempts multiple common access
        patterns to extract victory points and common counts (settlements,
        cities, roads). If extraction fails, the evaluator falls back to 0.

        Args:
            game (Game): complete game state. read-only. Use game.copy() to
                         create hypothetical states.
            playable_actions (Iterable[Action]): legal options for this turn.
        Returns:
            action: chosen element of playable_actions, or None if no options.
        """
        # Defensive: if no actions available, return None
        if not playable_actions:
            if DEBUG:
                print('FooPlayer.decide: no playable_actions -> returning None')
            return None

        # Convert playable_actions to a list so we can sample and index
        try:
            actions = list(playable_actions)
        except Exception:
            # If iterable cannot be converted, fall back to returning first
            if DEBUG:
                print('FooPlayer.decide: playable_actions not list-like; defaulting to first')
            try:
                return playable_actions[0]
            except Exception:
                return None

        # Sample candidate actions if there are too many
        if len(actions) > MAX_ACTIONS_TO_EVALUATE:
            candidates = random.sample(actions, MAX_ACTIONS_TO_EVALUATE)
            if DEBUG:
                print(f'FooPlayer.decide: sampled {len(candidates)} of {len(actions)} actions to evaluate')
        else:
            candidates = actions
            if DEBUG:
                print(f'FooPlayer.decide: evaluating all {len(candidates)} actions')

        # Evaluate each candidate action by applying it to a copy of the game
        scores = []  # list of (action, score)
        for i, action in enumerate(candidates):
            try:
                # Copy the game to avoid mutating the original
                new_game = game.copy()

                # Apply the candidate action on the copied game.
                # The standard Game API exposes execute(action) to apply an action.
                # We try both .execute and .apply for defensive compatibility.
                executed = False
                try:
                    new_game.execute(action)
                    executed = True
                except Exception:
                    # Some versions may expose a differently named method.
                    try:
                        new_game.apply(action)
                        executed = True
                    except Exception:
                        executed = False

                if not executed:
                    # If we couldn't apply the action on the copy, mark it as
                    # very poor and continue.
                    if DEBUG:
                        print(f'FooPlayer.decide: failed to execute candidate action {i}; marking score -inf')
                    scores.append((action, float('-inf')))
                    continue

                # Evaluate the successor state
                score = self._evaluate_state(new_game)
                scores.append((action, score))
                if DEBUG:
                    print(f'FooPlayer.decide: action #{i} -> score {score}')

            except Exception as e:
                # Catch-all: do not let the player crash the harness. Penalize
                # the action and continue evaluating others.
                if DEBUG:
                    print(f'FooPlayer.decide: exception while evaluating action #{i}: {e}! Marking -inf')
                scores.append((action, float('-inf')))

        # Choose the best action. If all are -inf or evaluation failed, fall back
        # to the original first-action policy.
        if not scores:
            if DEBUG:
                print('FooPlayer.decide: no scores produced -> defaulting to first action')
            return actions[0]

        # Compute the maximum score
        max_score = max(score for (_, score) in scores)
        # Filter all actions that have the max score (handle ties)
        best_candidates = [a for (a, s) in scores if s == max_score]

        if not best_candidates or max_score == float('-inf'):
            # All evaluations failed; fallback
            if DEBUG:
                print('FooPlayer.decide: all evaluations failed -> defaulting to first action')
            return actions[0]

        chosen = random.choice(best_candidates)
        if DEBUG:
            try:
                # Try to pretty-print a small summary for debugging
                print(f'FooPlayer.decide: selected action -> {repr(chosen)} with score {max_score}')
            except Exception:
                print('FooPlayer.decide: selected an action (repr failed)')

        return chosen

    def _evaluate_state(self, game):
        """Static evaluation of a game state from this player's perspective.

        The evaluator attempts multiple common access patterns to extract
        victory points and simple progress indicators (settlements, cities,
        roads). The returned score is primarily the victory points (higher is
        better). Secondary counts are used as small tiebreakers.

        This function is defensive to avoid attribute errors across different
        engine versions.

        Returns:
            float: heuristic score for the state (larger is better)
        """
        color = self.color
        vp = None
        settlements = None
        cities = None
        roads = None

        # Try a number of plausible attribute access patterns. Use try/except
        # blocks liberally because different engine versions expose different
        # structures.
        try:
            players = game.state.players
        except Exception:
            players = None

        # Attempt to access player state by Color key
        player_state = None
        if players is not None:
            try:
                player_state = players[color]
            except Exception:
                # Maybe players is a list keyed by integer colors
                try:
                    idx = int(color)
                    player_state = players[idx]
                except Exception:
                    player_state = None

        # Extract victory points with common attribute names
        if player_state is not None:
            for attr in ('victory_points', 'victoryPoints', 'vp', 'points'):
                try:
                    val = getattr(player_state, attr)
                    # If it's a callable (method), call it
                    if callable(val):
                        val = val()
                    vp = int(val)
                    break
                except Exception:
                    vp = None

            # Try dictionary-style if attributes failed
            if vp is None:
                try:
                    if isinstance(player_state, dict):
                        for key in ('victory_points', 'vp', 'points'):
                            if key in player_state:
                                vp = int(player_state[key])
                                break
                except Exception:
                    vp = None

            # Extract simple asset counts to break ties
            for attr in ('settlements', 'settle_count', 'settlement_count', 'settles'):
                try:
                    val = getattr(player_state, attr)
                    if callable(val):
                        val = val()
                    settlements = int(val)
                    break
                except Exception:
                    settlements = None

            for attr in ('cities', 'city_count'):
                try:
                    val = getattr(player_state, attr)
                    if callable(val):
                        val = val()
                    cities = int(val)
                    break
                except Exception:
                    cities = None

            for attr in ('roads', 'road_count'):
                try:
                    val = getattr(player_state, attr)
                    if callable(val):
                        val = val()
                    roads = int(val)
                    break
                except Exception:
                    roads = None

        # Fallbacks if extraction failed: try to compute from visible board pieces
        # (e.g., lengths of lists). This is optional and best-effort.
        if vp is None and players is not None:
            try:
                # If player_state contains lists of pieces, inspect lengths
                if isinstance(player_state, dict):
                    # Look for settlement/city lists
                    s = None
                    for key in ('settlements', 'settle_list'):
                        if key in player_state and isinstance(player_state[key], (list, tuple)):
                            s = len(player_state[key])
                            break
                    if s is not None:
                        settlements = settlements or s
                # We intentionally do not try to derive vp from the board in a
                # brittle way; leave vp as None and fall back to 0.
            except Exception:
                pass

        # Final fallback: if we couldn't determine vp, set to 0
        if vp is None:
            vp = 0

        # Build a composite score. Main contributor is victory points. Add
        # small weighted bonuses for settlements/cities/roads if available.
        score = float(vp)
        if settlements is not None:
            score += 0.01 * float(settlements)
        if cities is not None:
            score += 0.02 * float(cities)
        if roads is not None:
            score += 0.005 * float(roads)

        return score
\end{python}

\end{document}

%% file: math_commands.tex
\usepackage{amsmath,amsfonts,bm}

\def\eqref#1{equation~\ref{#1}}
\def\1{\bm{1}}

\DeclareMathAlphabet{\mathsfit}{\encodingdefault}{\sfdefault}{m}{sl}
\SetMathAlphabet{\mathsfit}{bold}{\encodingdefault}{\sfdefault}{bx}{n}